%% file: main.tex
\documentclass{article}
\usepackage[T1]{fontenc}
\usepackage[table]{xcolor}
\PassOptionsToPackage{hypertexnames=false}{hyperref}
\usepackage[final]{corl_2026}
\usepackage[utf8]{inputenc}
\usepackage{amsthm,amsmath,amssymb}
\usepackage{graphicx,multirow,booktabs}
\usepackage{algorithm,algorithmic}
\usepackage{tikz}
\usetikzlibrary{arrows.meta,positioning,calc}
\newcommand{\MERL}{\textsc{MeRL}}

\newcommand{\crproselayout}{\setlength{\parfillskip}{0pt plus 0.30\linewidth}}
\makeatletter
\let\croriginalmakecaption\@makecaption
\long\def\@makecaption#1#2{{\crproselayout\croriginalmakecaption{#1}{#2}}}
\makeatother
\makeatletter
\renewcommand{\@conferencelocation}{Austin, Texas, USA}
\makeatother
\hypersetup{
  pdftitle={Trust-Calibrated VLA Policy Post-Training with Evolving Imagination},
  pdfauthor={Wenzhan Li, Yiran Qin, Heng Zhou, Huirui Wang, Yulan Guo, Ruimao Zhang}
}
\title{Trust-Calibrated VLA Policy Post-Training \\ with Evolving Imagination}
\author{
  Wenzhan Li$^{1,2}$ \quad Yiran Qin$^{3,4,5}$ \quad Heng Zhou$^{4,6}$ \quad
  Huirui Wang$^{1}$ \\ \quad \textbf{Yulan Guo}$^{1,2}$ \quad \textbf{Ruimao Zhang}$^{1}$\thanks{Corresponding author.} \\
  $^1$Sun Yat-sen University \quad $^2$Shenzhen Loop Area Institute \\ $^3$The Chinese University of Hong Kong, Shenzhen \quad $^4$Shanghai AI Laboratory \\ $^5$University of Oxford \quad $^6$University of Science and Technology of China
}
\begin{document}
\maketitle
\input{sec/0_abs_0523}
\crproselayout
\input{sec/1_intro_0526}
\input{sec/2_related_0523}
\input{sec/4_method_0513}
\input{sec/5_exp_0528}
\input{sec/888_limit}
\clearpage
\setlength{\parfillskip}{0pt plus 1fil}
\bibliography{example}
\clearpage
\crproselayout
\input{sec/999_appendix}
\end{document}

%% file: sec/0_abs_0523.tex
\begin{abstract}
Vision-Language-Action (VLA) models are often pretrained via imitation learning, making reinforcement learning (RL) post-training important when deployment moves beyond demonstration coverage.
Existing refinement paradigms expose a reliability--efficiency tension: model-free refinement avoids simulator bias but is interaction-intensive, whereas model-based refinement can exploit unreliable imagination even as its simulator adapts.
We propose \textbf{Model-Evolving Reinforcement Learning (\MERL{})}, a trust-calibrated framework that couples policy optimization with an online-updated simulator comprising a world model and a reward proxy head.
Across stages, simulator reliability controls the imagined-data ratio and rollout horizon; within rollouts, chunk-level trust controls replay selection and optimization weight.
For recursive imagination without exact grounded continuations, a residual predictor estimates trust without querying future ground truth.
On perturbed LIBERO-PRO tasks, \MERL{} reaches $79.7\%$ average success, improves high-threshold learning speed by $10.9\%$ over the matched model-free control, and outperforms same-protocol external baselines across three training seeds.
Fixed-data adaptation from 50 demonstrations per task achieves $80.0\%$ average success on two physical-robot tasks.
\end{abstract}
\keywords{World Models, Reinforcement Learning, VLA Policy Post-Training}

%% file: sec/1_intro_0526.tex
\section{Introduction}
\label{sec:introduction}

Vision-Language-Action (VLA) models have become a scalable paradigm for language-conditioned robot manipulation, mapping visual observations and language instructions directly to executable actions~\cite{brohan2022rt,zitkovich2023rt,openvla,team2024octo,black2024pi_0}.
Yet most VLA policies still rely heavily on imitation learning, leaving their robustness tightly bounded by expert demonstration coverage.
When deployment moves beyond this coverage, errors accumulate, recovery remains weak, and adaptation to unseen perturbations is limited.
Reinforcement learning (RL) therefore offers a natural post-training route, allowing policies to improve through interaction rather than remain constrained by demonstrations.

However, applying RL to VLA post-training reveals a progression of bottlenecks as methods move from real interaction toward imagined experience.
Model-free refinement updates policies directly from real interaction and therefore avoids simulator bias, but it is costly, sample-inefficient, and often unstable in long-horizon, high-dimensional manipulation settings~\cite{liu2025can,lu2025vla,li2025simplevla,chen2025pi_,zang2025rlinf}.
Static model-based refinement reduces interaction cost by training policies with imagined rollouts from learned simulators~\cite{li2025vla,xiao2025world,jiang2025world4rl,zhu2025wmpo,zhang2025prophrl,intelligence2026pi}, but its simulator can become stale as the policy-induced distribution shifts.
Online-updated model-based refinement further adapts the simulator during policy improvement to better match newly induced states and environments~\cite{yang2026rise,liu2026world,jiang2026wovr}.
Simulator adaptation addresses staleness, while selective model usage and reliability-aware VLA refinement also study how imagined data should influence learning~\cite{pan2020masked,huang2021reweight,jiang2026wovr}.
A remaining question is how one reliability interface can jointly govern the stage-level use and chunk-level influence of visual imagination as the policy and simulator adapt.
This auditing gap becomes fundamental because simulator reliability is neither stationary nor homogeneous; it changes over training and varies across imagined rollouts.
This raises two central questions: \textit{how should VLA refinement estimate the trustworthiness of simulator-generated imagination as the simulator evolves, and how should this trust govern which imagined experience enters policy optimization?}

\begin{figure}[t!]
\centering
\includegraphics[width=1.0\textwidth]{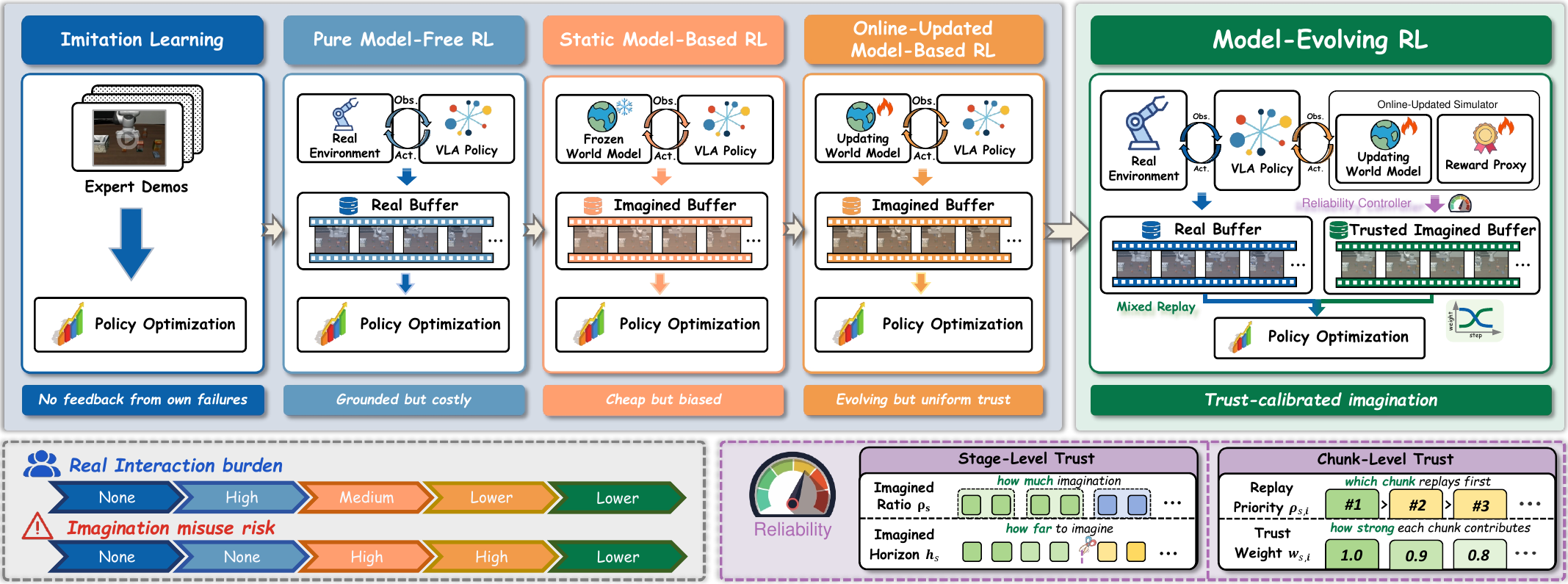}

\caption{
\textbf{Paradigm evolution toward trust calibration.}
The columns illustrate design patterns represented by our matched refinement controls, from imitation and model-free RL to static and online-updated model-based RL.
\MERL{} uses simulator reliability to schedule, select, and weight rollouts before the subsequent mixed real--imagined policy optimization.
}
\label{fig:merl_framework}

\end{figure}

Fig.~\ref{fig:merl_framework} illustrates the trust-calibrated mechanism introduced by \MERL{}.
To produce imagination, \MERL{} uses an evolving simulator composed of a world model and a reward proxy head, which provide short-horizon visual futures and local progress estimates for policy optimization.
The main contribution is not the simulator architecture itself, but the trust layer placed between simulator-generated imagination and policy learning.
As the simulator evolves, \MERL{} estimates trust at the level of each online refinement cycle through stage-level trust.
This global reliability estimate is converted into scheduling signals that regulate the strength and temporal extent of simulator-generated imagination.
During policy optimization, \MERL{} applies trust at the level of generated rollout segments through chunk-level trust.
Each imagined chunk is locally assessed, and its reliability determines replay priority and optimization weight.
This hierarchy specifies how trust is estimated from an evolving simulator and how it is operationalized before imagined experience shapes the policy.

The core insight behind \MERL{} is that simulator evolution and simulator trust are distinct problems.
Updating the simulator can make imagination more relevant to the current policy, but it does not guarantee that every generated rollout is equally useful for optimization.
\MERL{} therefore treats trust as an auditing interface between simulator-generated imagination and policy learning.
Stage-level trust prevents premature overuse of unreliable imagination, while chunk-level trust suppresses locally biased samples even when average simulator quality improves.
This design keeps the update close to grounded refinement when the simulator is weak and increases the use of imagination as its estimated reliability improves.
Fig.~\ref{fig:merl_framework} summarizes these design axes. Our matched online-updated control retains simulator adaptation while removing trust scheduling and chunk control, enabling a direct test of their contribution.

Our contributions are three-fold:
\textbf{(1)} We identify the reliability bottleneck of evolving simulators in online VLA post-training, and formulate it as a \emph{trust-calibration} problem where imagined rollouts should influence policy learning according to their estimated reliability.
\textbf{(2)} We propose \textbf{\MERL{}}, a trust-calibrated model-evolving reinforcement learning framework that combines online simulator evolution with two-level trust control, enabling adaptive use of grounded policy updates and simulator-generated imagination.
\textbf{(3)} We provide controlled ablations, same-protocol external comparisons, no-oracle trust diagnostics, and fixed-data physical deployment. On perturbed LIBERO-PRO tasks, \MERL{} reduces high-threshold refinement stages over the matched MFRL control by \textbf{10.9\%} and improves final success over the matched Online-Updated MBRL control by \textbf{3.5} points.

%% file: sec/2_related_0523.tex
\section{Related Work}
\label{sec:related_work}

\textbf{Vision-Language-Action Policies.}
Vision-Language-Action (VLA) policies map visual observations and language instructions to executable robot actions, and have become a scalable paradigm for language-conditioned manipulation.
Early systems formulate control as autoregressive action generation conditioned on vision and language~\cite{brohan2022rt,zhao2023learning}, while later works build on pretrained vision-language backbones and richer action representations to improve semantic grounding, transferability, and generalist control~\cite{zitkovich2023rt,openvla,team2024octo,black2024pi_0,kim2025fine,intelligence2025pi_,zhang2025ta}.
Despite rapid progress in architectures and training data, most VLA policies are still pretrained and adapted mainly through supervised imitation.
As a result, policy robustness remains bounded by demonstration coverage, leaving adaptation to unseen perturbations, dynamics shifts, and failure states insufficient.

\textbf{Model-Free Reinforcement Learning for VLA Post-Training.}
To overcome the limitations of pure imitation learning, a growing line of work studies model-free reinforcement learning (MFRL) for post-training VLA policies.
Early efforts show that reinforcement learning can improve pretrained VLA policies through interaction~\cite{liu2025can}, while subsequent works scale this paradigm to long-horizon and multi-task settings with trajectory-level optimization, distributed training, and VLA-specific rollout pipelines~\cite{lu2025vla,li2025simplevla}.
More recent studies further explore online fine-tuning for flow-based VLA models, unified RL infrastructures, continual test-time adaptation, and stronger long-horizon policy improvement~\cite{chen2025pi_,zang2025rlinf,bai2025evolve,intelligence2025pi,li2025gr}.
These works demonstrate that reinforcement learning can substantially improve VLA policies beyond supervised fine-tuning.
However, MFRL still depends heavily on costly real or simulator interaction, and its performance is often constrained by sample inefficiency, reward sensitivity, and unstable long-horizon optimization.

\textbf{Model-Based Reinforcement Learning for VLA Post-Training.}
Complementary to MFRL, model-based reinforcement learning (MBRL) reduces interaction cost by learning a simulator and refining policies with imagined experience.
In VLA post-training, recent methods instantiate this simulator as a visual world model for policy refinement.
Static MBRL uses a pretrained or separately trained simulator to generate imagined rollouts~\cite{li2025vla,xiao2025world,jiang2025world4rl,zhu2025wmpo,zhang2025prophrl,intelligence2026pi}.
This reduces real-interaction burden, but imagined data can become stale or biased as the policy-induced distribution shifts.
Online-updated MBRL further adapts the simulator during policy improvement to better match newly induced states and environments~\cite{yang2026rise,liu2026world,jiang2026wovr}.
Selective model usage is established in classical MBRL: MBPO uses short branched rollouts~\cite{janner2019trust}, Pan et al.\ mask transitions using ensemble uncertainty~\cite{pan2020masked}, and Huang et al.\ learn transition weights from real-loss meta-gradients~\cite{huang2021reweight}.
WoVR also studies reliable world-model simulation and policy--model co-evolution for VLA post-training~\cite{jiang2026wovr}.
Our contribution is the coupling of stage-level ratio/horizon scheduling with chunk-level replay/gradient control through visual and proxy residual estimates, including no-oracle scoring of recursive VLA imagination.
Online simulator updating supplies the adaptive model; the two-level reliability-to-policy interface is the focus of \MERL{}.

%% file: sec/4_method_0513.tex
\section{Method}
\label{sec:method}

\begin{figure}[tphb]
    \centering
    \includegraphics[width=\linewidth]{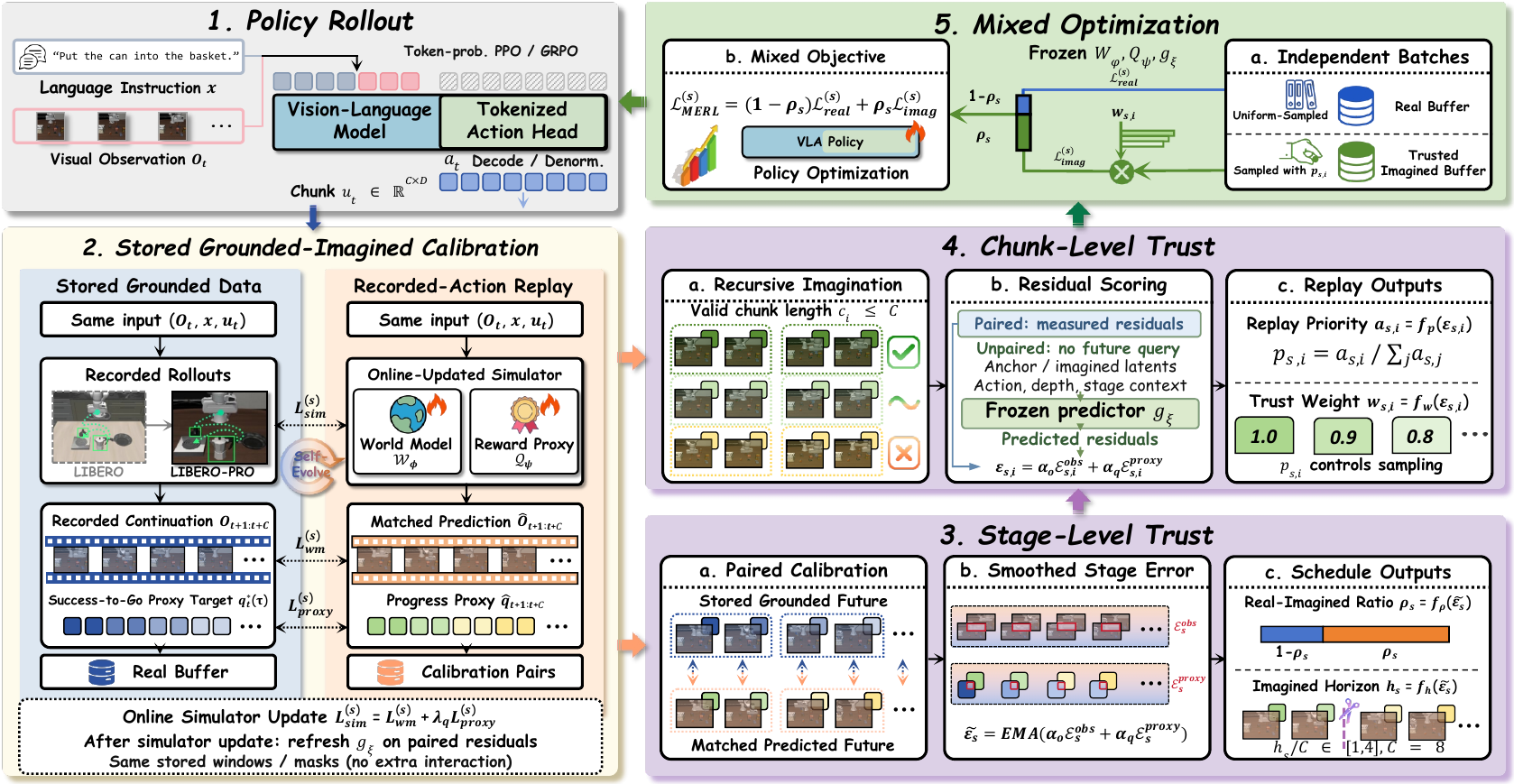}
    
    \caption{
    \textbf{Overview of the \MERL{} pipeline.}
    At each online stage $s$, the policy collects grounded rollouts and updates an evolving simulator composed of a world model $\mathcal{W}_{\phi}$ and a reward proxy head $\mathcal{Q}_{\psi}$.
    Calibration reuses stored grounded continuations with the same anchors and recorded actions; stage-level trust maps their measured errors to $\rho_s$ and $h_s$.
    A residual predictor refreshed after simulator update supplies no-oracle scores for unpaired recursive chunks and is frozen during imagination and policy optimization.
    Chunk-level trust then evaluates generated imagined chunks, assigns replay probabilities $p_{s,i}$ and optimization weights $w_{s,i}$, and combines trusted imagined chunks with grounded rollouts for mixed policy optimization.
    }
  \label{fig:merl_pipeline}
  
\end{figure}

Fig.~\ref{fig:merl_pipeline} summarizes the refinement cycle: grounded data update the simulator, stage-level trust schedules the imagined ratio and horizon, and chunk-level trust controls replay and optimization.
We formalize these components below under sparse terminal success supervision.

\subsection{Problem Formulation}
\label{sec:problem_formulation}

We consider online post-training of a pretrained Vision-Language-Action (VLA) policy $\pi_\theta$ in a perturbed manipulation environment.
Training proceeds in online stages indexed by $s$.
Each stage denotes one refinement cycle, where the current policy collects grounded rollouts, the simulator is updated and evaluated, imagined chunks are generated with a scheduled ratio and horizon, and the policy is optimized on mixed real--imagined experience.

Let $t$ denote an anchor time at which the policy issues a new action chunk.
Given a visual observation $\mathbf{o}_t\in\mathcal{O}$ and task instruction $\mathbf{x}\in\mathcal{X}$, the policy samples categorical action tokens $\mathbf{a}_t\sim\pi_\theta(\cdot\mid\mathbf{o}_t,\mathbf{x})$ and decodes/denormalizes them to $\mathbf{u}_t\in\mathbb{R}^{C\times D}$.
Here $C$ is the action chunk length and $D$ is the dimension of each environment-step action; policy ratios use token probabilities, while the environment and simulator consume continuous commands.
For a relative offset $j$, $\mathbf{o}_{t+j}$ is the post-transition observation; $\mathbf{u}_{t+j}$ denotes the aligned low-level action input, with the chunk context distinguishing it from the full $\mathbf{u}_t$.
Executing the policy in the real environment induces a trajectory $\tau$ with a sparse terminal success label $z(\tau)\in\{0,1\}$.

At stage $s$, the grounded batch records action chunks and observed continuations:
\[
\mathcal{B}^{\mathrm{real}}_s=\{(\mathbf{o}_t,\mathbf{x},\mathbf{u}_t,\mathbf{o}_{t+1:t+C},z(\tau))\},
\]
where $\mathbf{o}_{t+1:t+C}$ is produced by real environment interaction under $\mathbf{u}_t$.
The imagined batch is
\[
\mathcal{B}^{\mathrm{imag}}_s=\{(\tilde{\mathbf{o}}_t,\mathbf{x},\tilde{\mathbf{u}}_{t:t+h_s-1},\hat{\mathbf{o}}_{t+1:t+h_s},\hat{\mathbf{q}}_{t+1:t+h_s})\},
\]
where $\tilde{\mathbf{o}}_t$ is an anchor observation sampled for imagination, $\tilde{\mathbf{u}}_{t:t+h_s-1}$ is the action sequence obtained by rolling out the current policy inside the simulator, $\hat{\mathbf{o}}$ denotes simulator-predicted observations, and $\hat{\mathbf{q}}$ denotes proxy progress scores along the imagined continuation.
The scheduled horizon satisfies $1\le h_s\le H_{\max}$, where $h_s$ is measured in environment steps and $H_{\max}$ is the maximum allowed imagined rollout horizon.
Thus, $C$ counts actions per policy call, whereas $h_s$ counts imagined environment steps.
When $h_s>C$, the policy is queried recursively inside the simulator and the resulting action chunks are unrolled into $\tilde{\mathbf{u}}_{t:t+h_s-1}$.

We optimize $\pi_\theta$ on the two batches while controlling bias from imperfect simulator predictions.

\subsection{Online Simulator Evolution}
\label{sec:online_world_model_evolution}

\MERL{} instantiates the online-updated simulator with a world model $\mathcal{W}_\phi$ and a reward proxy head $\mathcal{Q}_\psi$.
We reserve \emph{world model} for the video predictor and use \emph{evolving} to denote online adaptation of the composite simulator.
At stage $s$, both components are updated from grounded trajectories collected by the current policy.
The simulator then outputs two signals for imagination: predicted visual continuations from $\mathcal{W}_\phi$ and local proxy progress scores from $\mathcal{Q}_\psi$.
This online update makes the simulator better aligned with the current policy-induced distribution than a frozen simulator, but it does not by itself guarantee that all generated rollouts are reliable.

Given a visual observation $\mathbf{o}_t$, instruction $\mathbf{x}$, and action chunk $\mathbf{u}_t$, the world model predicts one chunk of future observations as $\hat{\mathbf{o}}_{t+1:t+C}=\mathcal{W}_\phi(\mathbf{o}_t,\mathbf{x},\mathbf{u}_t)$, where $C$ is the action chunk length.
When the scheduled imagined horizon satisfies $h_s>C$, the policy and world model are recursively queried inside the simulator until the rollout reaches $h_s$ environment steps.
For each predicted observation-action pair, the reward proxy produces a bounded progress score $\hat{q}_{t+j}=\mathcal{Q}_\psi(\hat{\mathbf{o}}_{t+j},\mathbf{u}_{t+j})\in[0,1]$.
Thus, an imagined rollout provides both a visual sequence $\hat{\mathbf{o}}_{t+1:t+h_s}$ and a proxy-score sequence $\hat{\mathbf{q}}_{t+1:t+h_s}$ for policy optimization.
Implementation follows App.~\ref{app:simulator_architecture}.

The real environment provides only a sparse terminal success label $z(\tau)\in\{0,1\}$, which does not directly supervise intermediate imagined progress.
We therefore train the proxy head with a truncated success-to-go target.
For a trajectory $\tau$ of length $T$, define
\begin{equation}
q_t^{\star}(\tau)
=
z(\tau)\,\gamma_q^{\,\min(T-1-t,\;H_q-1)},
\label{eq:proxy_target_method}
\end{equation}
where $\gamma_q\in(0,1]$ is a discount factor and $H_q$ is a truncation horizon.
This target lies in $[0,1]$: for successful trajectories, states closer to completion receive larger values, while failed trajectories remain zero.
The target preserves terminal success and supervises local progress for proxy training.

The simulator is trained with $\mathcal{L}_{\mathrm{sim}}^{(s)}=\mathcal{L}_{\mathrm{wm}}^{(s)}+\lambda_q\mathcal{L}_{\mathrm{proxy}}^{(s)}$, where $\mathcal{L}_{\mathrm{wm}}^{(s)}$ supervises visual prediction, $\mathcal{L}_{\mathrm{proxy}}^{(s)}$ supervises the proxy head with $q_t^\star(\tau)$, and $\lambda_q$ controls the proxy-loss weight.
Detailed objectives are provided in App.~\ref{app:reward_proxy_details}.
Since visual prediction errors can still induce biased proxy scores on imagined observations, the next section estimates simulator reliability before deciding the imagined-data ratio and rollout horizon.

\subsection{Trust-Calibrated Imagination Scheduling}
\label{sec:trust_calibrated_scheduling}

Stage-level trust determines how strongly and how far \MERL{} uses simulator-generated imagination at training stage $s$.
Its input is the simulator mismatch measured after online simulator update, and its outputs are the imagined-data ratio $\rho_s$ and rollout horizon $h_s$ for the subsequent policy update.
The need to schedule both quantities comes from a simple fact: proxy supervision on imagined observations inherits visual prediction errors from the world model.

\paragraph{Simulator-induced proxy bias.}
Consider an anchor time $t$ and an imagined continuation of horizon $h\le h_s$.
Let $\hat{q}_{t}^{(h)}$ denote the discounted aggregate of proxy scores evaluated on simulator-predicted observations $\{\hat{\mathbf{o}}_{t+j}\}_{j=1}^{h}$, and let $q_{t}^{(h)}$ denote the corresponding analytical aggregate evaluated on grounded observations $\{\mathbf{o}_{t+j}\}_{j=1}^{h}$ under the same action continuation.
The grounded aggregate is not used as an online training signal; it only measures the bias introduced when the proxy head is evaluated on imagined observations.
We define the pointwise visual prediction discrepancy as $\delta_{s,t,j}=\Delta(\hat{\mathbf{o}}_{t+j},\mathbf{o}_{t+j})$, where $\Delta(\cdot,\cdot)$ measures observation-feature discrepancy.

\textbf{Proposition 1.}
If $\mathcal{Q}_{\psi}$ is $L_q$-Lipschitz with respect to observation features when the action input is fixed, then for an imagined continuation of horizon $h$,
\begin{equation}
\left|
\hat{q}_{t}^{(h)} - q_{t}^{(h)}
\right|
\le
L_q
\sum_{j=1}^{h}
\gamma_q^{j-1}\delta_{s,t,j}.
\label{eq:proxy_bias_bound_method}
\end{equation}
This bound shows that imagined proxy bias accumulates from visual prediction discrepancies along the rollout.
Thus, even a well-supervised proxy head can produce biased progress estimates when evaluated on inaccurate imagined observations, especially as the horizon grows.
This motivates scheduling the imagined horizon according to simulator reliability rather than using a fixed rollout length.
Formal definitions and proof are provided in App.~\ref{app:proofs}.

\paragraph{Stage-level reliability estimate.}
After updating the simulator, \MERL{} estimates a stage-level reliability error $\bar{\varepsilon}_s=\alpha_o\mathcal{E}^{\mathrm{obs}}_s+\alpha_q\mathcal{E}^{\mathrm{proxy}}_s$ from stored grounded windows.
The simulator replays each recorded action continuation from the same anchor and instruction; this calibration adds no environment interaction.
Here $\mathcal{E}^{\mathrm{obs}}_s$ measures short-horizon visual prediction mismatch, $\mathcal{E}^{\mathrm{proxy}}_s$ measures proxy-progress mismatch against the success-to-go target, and $\alpha_o,\alpha_q$ are reliability-aggregation weights.
To reduce online fluctuations, \MERL{} uses an exponentially smoothed error $\tilde{\varepsilon}_s$ as the global control signal.
Concrete definitions of these terms are given in App.~\ref{app:concrete_schedules}.

\paragraph{Scheduling rule.}
\MERL{} maps $\tilde{\varepsilon}_s$ to two coupled scheduling variables, $\rho_s=f_\rho(\tilde{\varepsilon}_s)$ and $h_s=f_h(\tilde{\varepsilon}_s)$, where both mappings are non-increasing with respect to $\tilde{\varepsilon}_s$.
The ratio $\rho_s\in[0,1]$ controls the contribution of imagined experience in policy optimization, while $h_s$ controls how far the simulator is rolled out from each anchor.
Thus, unreliable simulators induce smaller imagined ratios and shorter horizons, whereas more reliable simulators permit stronger and longer imagination.
Implementation-specific schedules are provided in App.~\ref{app:concrete_schedules}.

\paragraph{Effect of joint scheduling.}
The scheduler must control both the amount of imagined data and the rollout horizon because both factors scale the influence of simulator error.
Let $\Delta_s$ denote the aggregate discrepancy between imagined supervision and its grounded counterpart at stage $s$.
For scheduling analysis, let $\delta_{s,j}$ denote the stage-level visual prediction discrepancy at relative offset $j$, obtained by aggregating or upper-bounding $\delta_{s,t,j}$ over sampled anchors.

\textbf{Proposition 2.}
Suppose $\Delta_s\le K_\Delta\rho_s\sum_{j=1}^{h_s}\delta_{s,j}$ for some task-dependent constant $K_\Delta>0$.
If $\delta_{s,j}\le\tilde{\varepsilon}_s$ at every relative offset $j\le h_s$, the stage-level discrepancy satisfies
\begin{equation}
\Delta_s
\le
K_\Delta\,\rho_s\,h_s\,\tilde{\varepsilon}_s .
\label{eq:scheduling_bound}
\end{equation}
The measured average error is a scheduling statistic, not a certified upper bound on every rollout discrepancy; the proposition provides neither a convergence guarantee nor an empirical certificate.
Controlling only $\rho_s$ can still allow long biased rollouts, while controlling only $h_s$ can still admit too much unreliable imagined data.
We therefore couple the imagined-data ratio and rollout horizon through one shared stage-level trust signal (App.~\ref{app:proofs}).

\subsection{Trust-Aware Imagined Replay for Policy Optimization}
\label{sec:trust_aware_replay}

Stage-level trust controls the global use of imagination through the ratio $\rho_s$ and horizon $h_s$.
However, imagined chunks generated within the same stage can still have heterogeneous local reliability.
Chunk-level trust therefore determines which imagined chunks are more likely to enter policy optimization and how strongly they affect the update.

For each imagined chunk $i$ generated at stage $s$, \MERL{} estimates a chunk-level reliability error $\varepsilon_{s,i}=\alpha_o\mathcal{E}^{\mathrm{obs}}_{s,i}+\alpha_q\mathcal{E}^{\mathrm{proxy}}_{s,i}$, where $\mathcal{E}^{\mathrm{obs}}_{s,i}$ and $\mathcal{E}^{\mathrm{proxy}}_{s,i}$ denote estimated local observation mismatch and proxy-progress mismatch.
A smaller $\varepsilon_{s,i}$ indicates higher local trust.
\MERL{} converts this error into a normalized replay probability $p_{s,i}$ for sampling imagined chunks and an optimization weight $w_{s,i}$ for scaling their policy-gradient contribution.
Thus, $p_{s,i}$ controls local admission into the imagined batch, while $w_{s,i}$ controls local influence after sampling.
The exact priority normalization and weight functions are provided in App.~\ref{app:chunk_replay_details}.

\paragraph{Scoring without future ground truth.}
Exact grounded--predicted pairs use measured visual and proxy residuals under the same recorded action continuation.
For policy-generated chunks without such pairs, a residual predictor $g_\xi$ takes anchor/imagined latents, action, rollout depth, and stage context, and estimates $(\widehat{\mathcal{E}}^{\mathrm{obs}}_{s,i},\widehat{\mathcal{E}}^{\mathrm{proxy}}_{s,i})$.
We refresh $g_\xi$ on exact stored pairs after each simulator update, then freeze it during imagination and the ensuing policy update.
Recursive policy calls consume predicted observations; their chunk scores use only the predicted residuals, never a query to the future environment state.
These estimates enter the same weighted error $\varepsilon_{s,i}$ and the same $p_{s,i},w_{s,i}$ mappings as measured residuals.
App.~\ref{app:chunk_replay_details} specifies the information flow, and App.~\ref{app:no_oracle_validation} evaluates depths 2--4 with future observations withheld from scoring.

At stage $s$, \MERL{} samples grounded chunks into $\mathcal{B}^{\mathrm{real}}_s$ and imagined chunks into $\mathcal{B}^{\mathrm{imag}}_s$.
The two batches are sampled independently; grounded--imagined pairing is used only for reliability calibration when paired continuations are available, not for policy optimization.
Both branches use the same clipped policy-gradient form, while the imagined branch is additionally weighted by $w_{s,i}$.
The resulting real--imagined objective at stage $s$ is
\begin{equation}
\mathcal{L}_{\mathrm{MERL}}^{(s)}
=
(1-\rho_s)\mathcal{L}_{\mathrm{real}}^{(s)}
+
\rho_s\mathcal{L}_{\mathrm{imag}}^{(s)}.
\label{eq:merl_final_obj_slim}
\end{equation}
Here $\mathcal{L}_{\mathrm{real}}^{(s)}$ is computed on grounded rollouts, and $\mathcal{L}_{\mathrm{imag}}^{(s)}$ is computed on imagined rollouts sampled according to $p_{s,i}$ and weighted by $w_{s,i}$.
Therefore, $\rho_s$ controls the stage-level mixture between real and imagined branches, whereas $p_{s,i}$ and $w_{s,i}$ control the local admission and influence of individual imagined chunks.
The token-level branch objectives, validity masks, and complete online refinement procedure are provided in Apps.~\ref{app:policy_objective_details} and~\ref{app:full_algorithm}.

%% file: sec/5_exp_0528.tex
\section{Experiments}
\label{sec:experiments}

We evaluate whether \MERL{} improves policy post-training under deployment shifts, whether simulator updates and two-level trust provide complementary gains, and whether predicted trust is informative for recursive imagination and physical deployment.

\paragraph{Setup and metrics.}
We use LIBERO-PRO~\cite{zhou2025libero}, with Spatial, Object, Goal, and Long suites.
Task semantics remain fixed while deployment conditions shift, isolating adaptation from new-task acquisition.
Methods share OpenVLA-OFT initialization~\cite{kim2025fine}, action chunks, and grounded budgets.
A training stage is one outer refinement cycle, distinct from environment transitions or task horizon.
After stage $s$, each online method has $6s$ grounded trajectories under the same $\le3072s$ transition cap; calibration reuses stored windows and adds zero environment interaction.
We report final success rate (SR), normalized area under the SR learning curve (AUC), and stages to threshold (S2T).
S2T-H averages S2T@80 for Spatial/Object/Goal and S2T@65 for Long.
Threshold crossing is linearly interpolated between evaluated checkpoints, allowing fractional values.
Imagination and simulator updates incur model computation; S2T-H compares refinement speed under matched grounded budgets.
App.~\ref{app:exp_details} gives the metrics, budget accounting, and implementation.

\paragraph{Matched controls.}
We isolate components using the same policy optimizer and reward interface.
\textbf{SFT} evaluates the initialization without RL;
\textbf{MFRL} uses grounded refinement only;
\textbf{Static-MBRL} uses a frozen simulator, fixed ratio/horizon, and uniform replay;
\textbf{Static-MBRL+Trust} adds both trust levels to the frozen simulator;
\textbf{Online-Updated MBRL} updates the simulator while retaining fixed scheduling and uniform replay.
\textbf{\MERL{}} combines simulator updates and both trust levels.
These controls test components; concrete external algorithms are compared separately below.

\begin{table}[t]
\centering
\small
\setlength{\tabcolsep}{3pt}
\caption{Matched-control results on LIBERO-PRO. SR and AUC are on a percentage scale; S2T-H counts refinement stages. See Tab.~\ref{tab:external_results} for three-seed comparisons.}
\label{tab:main_results}
\begin{tabular}{lrrrrrrr}
\toprule
& \multicolumn{4}{c}{Suite SR (\%) $\uparrow$} & \multicolumn{3}{c}{Overall}\\
\cmidrule(lr){2-5}\cmidrule(lr){6-8}
Method & Spatial & Object & Goal & Long & Avg. SR $\uparrow$ & AUC $\uparrow$ & S2T-H $\downarrow$\\
\midrule
SFT & 38.6 & 42.4 & 23.8 & 15.9 & 30.2 & -- & --\\
MFRL & 80.4 & 83.1 & 79.4 & 66.0 & 77.2 & 67.9 & 103.4\\
Static-MBRL & 76.8 & 79.4 & 76.3 & 60.7 & 73.3 & 59.8 & 168.9\\
Static-MBRL+Trust & 78.2 & 80.8 & 78.4 & 62.8 & 75.1 & 63.5 & 145.8\\
Online-Updated MBRL & 79.5 & 82.0 & 77.0 & 66.1 & 76.2 & 65.4 & 139.6\\
\rowcolor{blue!8}
\MERL{} (Ours) & \textbf{82.8} & \textbf{85.2} & \textbf{81.9} & \textbf{68.8} & \textbf{79.7} & \textbf{70.6} & \textbf{92.1}\\
\bottomrule
\end{tabular}
\end{table}

\paragraph{Main results.}
Tab.~\ref{tab:main_results} shows gains in both final task performance and learning efficiency.
Relative to MFRL, \MERL{} improves Avg. SR from $77.2$ to $79.7$ and AUC from $67.9$ to $70.6$, while reducing S2T-H from $103.4$ to $92.1$ ($10.9\%$).
Static-MBRL underperforms MFRL, consistent with biased or stale imagination.
Adding trust improves the frozen-simulator control, but it remains below \MERL{}.
Online-Updated MBRL improves over Static-MBRL, yet \MERL{} raises its Avg. SR by $3.5$ points and reduces S2T-H from $139.6$ to $92.1$.
These comparisons support the complementary roles of simulator adaptation and reliability control.

\begin{table}[t]
\centering
\small
\setlength{\tabcolsep}{4pt}
\caption{Same-protocol external comparisons on LIBERO-PRO: mean $\pm$ standard deviation over three training seeds. $\dagger$ denotes a port changing benchmark/model assets while retaining the defining algorithm; each WoVR seed has an independent simulator.}
\label{tab:external_results}
\begin{tabular}{llccc}
\toprule
Method & Paradigm & Avg. SR $\uparrow$ & AUC $\uparrow$ & S2T-H $\downarrow$\\
\midrule
SimpleVLA-RL & Model-free & $77.1\pm1.1$ & $67.8\pm1.0$ & $104.1\pm7.0$\\
WMPO$^\dagger$ & Static model-based & $75.4\pm1.2$ & $63.2\pm1.3$ & $149.6\pm10.2$\\
WoVR$^\dagger$ & Online-updated & $77.8\pm0.9$ & $68.4\pm0.9$ & $101.3\pm6.2$\\
\rowcolor{blue!8}
\MERL{} & Trust-calibrated & $\mathbf{79.7\pm0.7}$ & $\mathbf{70.6\pm0.8}$ & $\mathbf{92.1\pm4.8}$\\
\bottomrule
\end{tabular}
\end{table}

\paragraph{External positioning and uncertainty.}
We compare SimpleVLA-RL~\cite{li2025simplevla}, WMPO~\cite{zhu2025wmpo}, and WoVR~\cite{jiang2026wovr}.
Methods in Tab.~\ref{tab:external_results} share SFT initialization, split, grounded budget, action chunk, optimizer/reward interface, checkpoints, and evaluation.
\MERL{} has the strongest mean SR/AUC/S2T-H among these methods.
Against WoVR, paired 95\% confidence intervals (CIs) give $\Delta\mathrm{AUC}=2.2\,[1.2,3.3]$ and $\Delta\mathrm{S2T\mbox{-}H}=-9.2\,[-15.4,-3.1]$.
Against the matched Online-Updated MBRL control, the three-seed contrast gives $\Delta\mathrm{AUC}=5.0\,[3.8,6.2]$.
External comparisons establish overall standing; the matched control isolates trust at a shared update cadence.

\paragraph{Where do the gains come from?}
The component ablation in Tab.~\ref{tab:ablation} retains simulator evolution while enabling either trust level alone.
Adding stage-level trust raises AUC from $65.4$ to $68.4$ and reduces S2T-H from $139.6$ to $113.2$.
Adding chunk-level trust raises AUC to $67.6$ and reduces S2T-H to $121.5$.
Using both gives the strongest Avg. SR, AUC, and S2T-H, consistent with complementary global scheduling and local replay control.
Simulator diagnostics and cross-suite results are in App.~\ref{app:extra_ablations}.
One-at-a-time $\{0.5,1,2\}\times$ sweeps of scheduler sharpness $\kappa$ and proxy-loss weight $\lambda_q$ change Avg. SR/AUC by at most $1.4/1.6$ points; every setting retains at least $+1.8/+3.1$ points over the matched online-updated control (App.~\ref{app:sensitivity}).

\begin{figure}[!t]
\centering
\includegraphics[width=\linewidth]{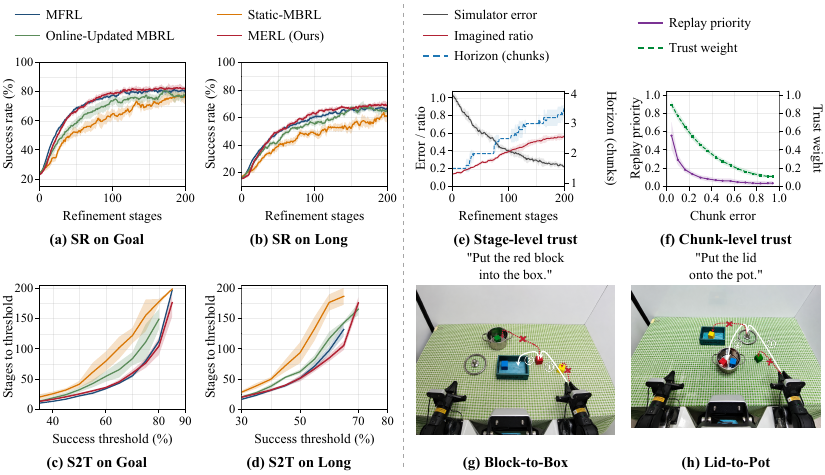}
\caption{Learning dynamics, two-level trust, and real-robot task setups. Left: SR and S2T on Goal and Long. Right: trust diagnostics and AgileX PiPER X setups for Block-to-Box and Lid-to-Pot. All training axes count outer refinement stages. Panel (e) displays the horizon in action-chunk units $h_s/C\in[1,4]$, with $C=8$; panel (f) displays replay priority and trust weight against chunk error.}
\label{fig:trust_dynamics}
\end{figure}

\paragraph{Learning dynamics and no-oracle trust.}
Fig.~\ref{fig:trust_dynamics} links policy improvement to the two trust controls.
Lower simulator error permits larger $\rho_s$ and $h_s$; lower chunk error receives larger replay probability and weight.
For unpaired recursive chunks, we evaluate the residual predictor with matched future observations withheld from scoring.
Across depths 2--4 ($N=1500$), Pearson $r=0.68$, Rank-AUC $=0.80$, ECE $=0.067$, and top-30\% low-error precision $=0.77$.
At depths 3--4 ($N=900$), the corresponding values are $0.64/0.77/0.074/0.74$, indicating useful but weaker prediction deeper into imagination.
Matched-count refinements from Top/Mid/Bottom trust bins achieve policy AUC $69.9\pm0.8/68.3\pm0.9/66.7\pm1.0$ over three seeds, with $\Delta_{\mathrm{Top-Bottom}}=3.2\,[2.0,4.4]$.
These results connect predicted trust to policy utility under this protocol.
Full depth-wise CIs and residual-predictor calibration diagnostics for these recursive rollouts appear in App.~\ref{app:no_oracle_validation}.

\begin{table}[t]
\centering
\small
\setlength{\tabcolsep}{6pt}
\caption{AgileX PiPER X closed-loop results: 20 trials/task after offline refinement from 50 demonstrations/task without extra physical training rollouts. Average-SR CIs are 95\% intervals.}
\label{tab:real_robot_success}
\begin{tabular}{lccc}
\toprule
Method & Block-to-Box & Lid-to-Pot & Avg. SR (\%) [95\% CI] $\uparrow$\\
\midrule
SFT & 8/20 & 6/20 & $35.0\,[22.1,50.5]$\\
MFRL & 12/20 & 10/20 & $55.0\,[39.8,69.3]$\\
Online-Updated MBRL & 13/20 & 11/20 & $60.0\,[44.6,73.7]$\\
\MERL{}-Visual & 16/20 & 15/20 & $77.5\,[62.5,87.7]$\\
\rowcolor{blue!8}
\MERL{} (full) & \textbf{17/20} & \textbf{15/20} & $\mathbf{80.0\,[65.2,89.5]}$\\
\bottomrule
\end{tabular}
\end{table}

\paragraph{Full-method physical deployment.}
Tab.~\ref{tab:real_robot_success} compares fixed-data adaptation on Block-to-Box and Lid-to-Pot under held-out placements and colors.
All methods share data, hardware interface, update budget, optimizer, and action horizon.
Online-Updated MBRL uses visual imagination; \MERL{}-Visual uses visual-only trust ($\alpha_q=0$); full \MERL{} enables $\mathcal{Q}_\psi$ and joint visual/proxy trust.
Full \MERL{} achieves the largest observed average SR ($80.0\%$), while \MERL{}-Visual obtains $77.5\%$.
Their CIs overlap, so these trials support full-method deployment without resolving a significant gain from enabling the proxy alone.
Training uses only the fixed demonstrations; physical evaluation is closed-loop inference.
App.~\ref{app:real_robot_eval} provides the full protocol, qualitative rollouts, and failure modes.

%% file: sec/888_limit.tex
\section{Conclusion and Limitations}
\label{sec:conclusion_limitations}

\paragraph{Conclusion.}
We presented \MERL{}, a trust-calibrated framework for VLA policy post-training with an online-updated simulator.
Stage-level scheduling and chunk-level replay control turn estimated simulator reliability into explicit policy-update decisions.
Controlled LIBERO-PRO experiments, same-protocol external comparisons, no-oracle trust diagnostics, and fixed-data physical deployment support the benefits of this coupling under the evaluated deployment shifts.

\paragraph{Limitations.}
Our validation covers tokenized VLA policies and structured tabletop tasks with fixed RGB sensing; LIBERO-PRO preserves task semantics and does not establish broad new-task learning or contact-rich generalization.
Trust is a predictive estimate based on visual and proxy residuals, and its accuracy decreases at deeper rollout depths; it does not certify physical correctness.
The trust-bin experiment evaluates predictive policy utility, while individual chunks' causal contributions remain unresolved.
Online simulator updates add computation, and calibration depends on grounded coverage even though it reuses stored data.
The physical study uses two tasks and offline refinement, with overlapping confidence intervals between full \MERL{} and \MERL{}-Visual; it does not establish online physical RL or a statistically resolved proxy-only gain.
Broader sensing, continuous-action generative policies, and larger closed-loop robot studies remain future work.

%% file: sec/999_appendix.tex
\appendix

\section{Full MERL Formulation}
\label{app:full_merl_formulation}

This appendix provides the full formulation of \MERL{} beyond the compact description in the main paper.
We first clarify notation, then detail the simulator architecture, simulator training objective, reliability schedules, chunk-level replay and weighting rules, policy objective, and online training algorithm.
All symbols are consistent with the main text.

\paragraph{Notation clarification.}
We distinguish four indices throughout the formulation.
The symbol $s$ indexes online training stages, where each stage corresponds to one outer-loop refinement cycle.
The symbol $t$ denotes an anchor time at which the policy issues an action chunk.
The symbol $j$ denotes a relative environment-step offset from the anchor time.
The symbols $h$ and $h_s$ denote imagined rollout horizons measured in environment steps.

The action chunk length is denoted by $C$, while the maximum allowed imagined rollout horizon is denoted by $H_{\max}$.
Thus, $C$ specifies how many low-level actions a single policy call outputs, whereas $h_s$ specifies how far the simulator is rolled out from an anchor at stage $s$, with $1\le h_s\le H_{\max}$.
When $h_s>C$, the policy and simulator are applied recursively inside the imagined rollout.
We use $\hat{\mathbf{o}}$ for simulator-predicted observations and $\tilde{\mathbf{o}}$ for anchor observations sampled for imagination.
For a relative offset $j$, $\delta_{s,t,j}$ denotes the pointwise visual prediction discrepancy between $\hat{\mathbf{o}}_{t+j}$ and $\mathbf{o}_{t+j}$ at stage $s$ and anchor time $t$.
Its stage-level aggregate or upper bound over sampled anchors is denoted by $\delta_{s,j}$.
In contrast, $\bar{\varepsilon}_s$ and $\tilde{\varepsilon}_s$ denote raw and smoothed stage-level reliability errors, while $\varepsilon_{s,i}$ denotes the estimated local reliability error of imagined chunk $i$.

For proxy notation, $q_t^\star(\tau)$ denotes the success-to-go proxy target constructed from the terminal success label of trajectory $\tau$.
The scalar $\hat{q}_{t+j}$ denotes a proxy prediction at relative offset $j$.
For a horizon $h$, $\hat{q}_{t}^{(h)}$ denotes the discounted proxy aggregate evaluated on simulator-predicted observations, while $q_{t}^{(h)}$ denotes the corresponding analytical aggregate evaluated on grounded observations under the same action continuation.
When discussing replay, $\widetilde{\mathcal{B}}^{\mathrm{imag}}_s$ denotes the candidate imagined chunks generated or retrieved at stage $s$, while $\mathcal{B}^{\mathrm{imag}}_s$ contains imagined chunks sampled from these candidates for subsequent policy optimization.

\subsection{Simulator Architecture}
\label{app:simulator_architecture}

\MERL{} instantiates the evolving simulator with two components: a controllable video-prediction world model $\mathcal{W}_\phi$ and a lightweight reward proxy head $\mathcal{Q}_\psi$.
The simulator takes an observation, instruction, and action input as conditions, and returns both predicted visual continuations and scalar proxy progress scores.
The world model supports visual imagination, while the proxy head converts sparse task success into local progress supervision that can be used during policy optimization.

\paragraph{World-model backbone.}
The world model follows controllable generative world models for robot manipulation~\cite{guo2025ctrl}.
Given a current observation $\mathbf{o}_t$, task instruction $\mathbf{x}$, and action chunk $\mathbf{u}_t\in\mathbb{R}^{C\times D}$, it predicts one chunk of future observations:
\begin{equation}
\hat{\mathbf{o}}_{t+1:t+C}
=
\mathcal{W}_{\phi}(\mathbf{o}_t,\mathbf{x},\mathbf{u}_t),
\label{eq:app_wm_arch_pred}
\end{equation}
where $C$ is the VLA action chunk length.
In implementation, RGB observations are encoded into visual latent tokens, and action chunks are projected into action-conditioning tokens.
The prediction backbone uses action-conditioned spatio-temporal transformers so that future visual tokens can attend to the corresponding action context.
The predicted sequence can be decoded into RGB frames for visualization or kept in latent space for reliability estimation and proxy evaluation.

The scheduled imagined horizon $h_s$ in the main text is not restricted to a single model call.
When $h_s>C$, \MERL{} recursively queries the policy and world model inside the simulator.
Specifically, the policy produces a new action chunk from the latest predicted observation, and $\mathcal{W}_\phi$ predicts the next visual chunk until the rollout reaches $h_s$ environment steps.
Thus, $C$ denotes the output length of one policy/world-model chunk, whereas $h_s$ denotes the scheduled simulator rollout horizon.

\paragraph{Reward proxy head.}
The reward proxy head $\mathcal{Q}_\psi$ maps observation-action latents to a bounded scalar progress score.
Given a grounded or predicted observation, a visual encoder produces observation tokens, and an action projection maps the aligned action input into action tokens.
The observation tokens attend to the action tokens through lightweight cross-attention blocks, and the fused representation is passed to an MLP head.
Its positive-class probability provides the proxy's bounded scalar estimate of local success-to-go progress in the interval $[0,1]$.

On grounded observations, the proxy evaluates the aligned observation-action pair as
\[
\hat{q}_{t+j}
=
\mathcal{Q}_{\psi}(\mathbf{o}_{t+j},\mathbf{u}_{t+j})
\in[0,1],
\]
On imagined observations, the same proxy instead evaluates the predicted pair as
\[
\hat{q}_{t+j}
=
\mathcal{Q}_{\psi}(\hat{\mathbf{o}}_{t+j},\mathbf{u}_{t+j})
\in[0,1].
\]
Here $\mathbf{u}_{t+j}$ denotes the action input aligned with the transition at relative offset $j$; it can be instantiated as a per-step action or a local action window depending on the implementation.
The proxy notation retains observation-action conditioning independently of action tokenization.

\paragraph{Role separation.}
The world model and reward proxy head serve different purposes.
The world model predicts what visual future may follow an action-conditioned rollout, while the proxy head estimates how much local task progress that transition represents.
This separation avoids forcing the video predictor to directly model sparse terminal success, and allows the progress estimator to be updated from success-to-go targets without changing the visual prediction objective.
It also provides two reliability signals: visual mismatch measures prediction fidelity, while proxy mismatch measures progress-supervision reliability.

\subsection{Detailed Simulator Objective}
\label{app:reward_proxy_details}

At training stage $s$, \MERL{} updates the simulator from grounded trajectories collected by the current policy.
The world model is supervised by one-chunk visual prediction, and the reward proxy head is supervised by a dense success-to-go target derived from sparse terminal success.

\paragraph{World-model objective.}
The grounded batch for one-chunk prediction is
\begin{equation}
\mathcal{B}^{\mathrm{real}}_s
=
\left\{
(\mathbf{o}_t,\mathbf{x},\mathbf{u}_t,\mathbf{o}_{t+1:t+C},z(\tau))
\right\},
\label{eq:app_real_batch}
\end{equation}
From each anchor and action chunk, the world model predicts $C$ future observations:
\begin{equation}
\hat{\mathbf{o}}_{t+1:t+C}
=
\mathcal{W}_{\phi}(\mathbf{o}_t,\mathbf{x},\mathbf{u}_t).
\label{eq:app_wm_pred}
\end{equation}
The world-model loss aggregates temporally weighted discrepancies over grounded chunks:
\begin{equation}
\mathcal{L}^{(s)}_{\mathrm{wm}}
=
\frac{1}{|\mathcal{B}^{\mathrm{real}}_s|}
\sum_{(\mathbf{o}_t,\mathbf{x},\mathbf{u}_t,\mathbf{o}_{t+1:t+C},z)\in\mathcal{B}^{\mathrm{real}}_s}
\sum_{j=1}^{C}
\omega_j
d(\hat{\mathbf{o}}_{t+j},\mathbf{o}_{t+j}),
\label{eq:app_wm_loss}
\end{equation}
where $d(\cdot,\cdot)$ denotes an observation discrepancy and $\omega_j\ge0$ is a temporal weighting coefficient.
Depending on the backbone, $d$ can be instantiated in pixel space, feature space, latent space, or as the reconstruction-equivalent term of a diffusion denoising objective.

\paragraph{Success-to-go proxy target.}
The environment provides a sparse terminal success label $z(\tau)\in\{0,1\}$.
To supervise local progress, \MERL{} defines a truncated discounted success-to-go target:
\begin{equation}
q_t^{\star}(\tau)
=
z(\tau)\,
\gamma_q^{\,\min(T-1-t,\;H_q-1)},
\label{eq:app_proxy_target}
\end{equation}
where $T$ is the trajectory length, $\gamma_q\in(0,1]$ is the proxy discount factor, and $H_q$ is the truncation horizon.
The target satisfies $q_t^\star(\tau)\in[0,1]$.
For successful trajectories, states closer to completion receive larger values; for failed trajectories, the target remains zero.
This target preserves the terminal task objective while providing bounded dense supervision for the proxy head.

\paragraph{Reward proxy objective.}
Grounded success-to-go supervision trains the proxy via
\begin{equation}
\mathcal{L}^{(s)}_{\mathrm{proxy}}
=
\frac{1}{|\mathcal{B}^{\mathrm{real}}_s|}
\sum_{(\mathbf{o}_t,\mathbf{x},\mathbf{u}_t,\mathbf{o}_{t+1:t+C},z)\in\mathcal{B}^{\mathrm{real}}_s}
\sum_{j=0}^{C-1}
\ell_q\!\left(
\mathcal{Q}_{\psi}(\mathbf{o}_{t+j},\mathbf{u}_{t+j}),
q^{\star}_{t+j}(\tau)
\right),
\label{eq:app_proxy_loss}
\end{equation}
where $\ell_q(\cdot,\cdot)$ is the proxy loss.
In our implementation, $\mathcal{Q}_\psi$ is trained as a binary progress classifier with soft target $q^\star_{t+j}(\tau)$, and its positive-class probability is used as the scalar proxy score.
With continuous progress labels, $\ell_q$ can instead use mean squared error or smooth $\ell_1$.

\paragraph{Joint simulator objective.}
The simulator is updated by
\begin{equation}
\mathcal{L}^{(s)}_{\mathrm{sim}}
=
\mathcal{L}^{(s)}_{\mathrm{wm}}
+
\lambda_q
\mathcal{L}^{(s)}_{\mathrm{proxy}},
\label{eq:app_sim_loss}
\end{equation}
where $\lambda_q\ge0$ controls the proxy-loss weight in simulator training.
This objective updates the visual predictor and the proxy estimator on grounded data from the current policy-induced distribution.
The training weight $\lambda_q$ is distinct from the reliability weights $\alpha_o,\alpha_q$ in App.~\ref{app:concrete_schedules}.

\subsection{Concrete Reliability Schedules}
\label{app:concrete_schedules}

The main paper defines stage-level trust through two monotone scheduling functions $f_\rho$ and $f_h$.
Here we give the concrete instantiation used in \MERL{}.
The scheduler takes a smoothed stage-level reliability error $\tilde{\varepsilon}_s$ as input and outputs the imagined-data ratio $\rho_s$ and rollout horizon $h_s$.

\paragraph{Stage-level reliability error.}
At training stage $s$, \MERL{} estimates simulator reliability from grounded calibration chunks collected by the current policy.
We reuse stored grounded windows and replay their recorded actions from the same anchor and instruction after simulator update.
The stored observations supply calibration targets without additional environment interaction.
Multi-chunk stored windows also supervise the residual predictor described in App.~\ref{app:chunk_replay_details}.
The raw stage-level reliability error combines visual prediction mismatch and proxy-progress mismatch:
\begin{equation}
\bar{\varepsilon}_s
=
\alpha_o\mathcal{E}^{\mathrm{obs}}_s
+
\alpha_q\mathcal{E}^{\mathrm{proxy}}_s,
\label{eq:app_stage_error}
\end{equation}
where $\alpha_o,\alpha_q\ge0$ are reliability-aggregation weights.
The observation mismatch is
\begin{equation}
\mathcal{E}^{\mathrm{obs}}_s
=
\frac{1}{|\mathcal{B}^{\mathrm{real}}_s|}
\sum_{(\cdot)\in\mathcal{B}^{\mathrm{real}}_s}
\sum_{j=1}^{C}
\omega_j
d(\hat{\mathbf{o}}_{t+j},\mathbf{o}_{t+j}),
\label{eq:app_obs_error}
\end{equation}
where $C$ is the action chunk length, $d(\cdot,\cdot)$ is the observation discrepancy, and $\omega_j\ge0$ is a temporal coefficient.
Proxy mismatch compares estimated progress and its target:
\begin{equation}
\mathcal{E}^{\mathrm{proxy}}_s
=
\frac{1}{|\mathcal{B}^{\mathrm{real}}_s|}
\sum_{(\cdot)\in\mathcal{B}^{\mathrm{real}}_s}
\sum_{j=0}^{C-1}
\nu_j
\left|
\mathcal{Q}_\psi(\mathbf{o}_{t+j},\mathbf{u}_{t+j})
-
q^{\star}_{t+j}(\tau)
\right|,
\label{eq:app_proxy_error}
\end{equation}
where $\nu_j\ge0$ is a temporal coefficient for proxy mismatch.
Thus, $\mathcal{E}^{\mathrm{obs}}_s$ measures how accurately the world model predicts grounded visual futures, while $\mathcal{E}^{\mathrm{proxy}}_s$ measures how accurately the reward proxy estimates local progress on grounded observation-action pairs.

\paragraph{Exponential smoothing.}
Online simulator updates can make the raw reliability error fluctuate across stages.
To avoid unstable schedule changes, \MERL{} uses an exponential moving average:
\begin{equation}
\tilde{\varepsilon}_s
=
\beta_{\varepsilon}\tilde{\varepsilon}_{s-1}
+
(1-\beta_{\varepsilon})\bar{\varepsilon}_s,
\qquad
\beta_{\varepsilon}\in[0,1).
\label{eq:app_error_smoothing}
\end{equation}
The scheduler interprets smaller $\tilde{\varepsilon}_s$ as higher stage-level simulator reliability.

\paragraph{Confidence score.}
We convert the smoothed error into a normalized confidence score:
\begin{equation}
\chi_s
=
\left(
1+
\left(
\frac{\tilde{\varepsilon}_s}{\varepsilon_{\mathrm{ref}}}
\right)^{\kappa}
\right)^{-1},
\qquad
\chi_s\in(0,1],
\label{eq:app_confidence_score}
\end{equation}
where $\varepsilon_{\mathrm{ref}}>0$ is a reference error scale and $\kappa>0$ controls the sharpness of the transition.
This inverse-power form maps lower reliability error to higher confidence.
The reference scale absorbs the units and normalization of the observation and proxy mismatch terms.

\paragraph{Imagination ratio schedule.}
The imagined-data ratio is
\begin{equation}
\rho_s
=
\rho_{\min}
+
(\rho_{\max}-\rho_{\min})\chi_s,
\qquad
0\le\rho_{\min}\le\rho_{\max}\le1.
\label{eq:app_ratio_schedule}
\end{equation}
Here $\rho_s$ controls the contribution of imagined experience in the mixed policy objective.
In all experiments, we set $\rho_{\min}=0.05$ and $\rho_{\max}=0.95$.
The lower bound keeps the update close to real-interaction refinement when the simulator is unreliable, while still retaining a small imagined branch for calibration and replay maintenance.
The upper bound makes the reliable-simulator regime close to model-based refinement, while retaining grounded experience for calibration.

\paragraph{Imagined horizon schedule.}
The imagined rollout horizon is
\begin{equation}
h_s
=
\left\lfloor
h_{\min}
+
(h_{\max}-h_{\min})\chi_s
\right\rfloor,
\qquad
1\le h_{\min}\le h_{\max}\le H_{\max}.
\label{eq:app_horizon_schedule}
\end{equation}
Here $h_s$ is measured in environment steps, and $H_{\max}$ is the maximum allowed imagined rollout horizon.
It should not be confused with the action chunk length $C$.
In our implementation, we set $h_{\min}=C$ and $h_{\max}=H_{\max}=4C$.
Since $C=8$, the scheduled horizon is 8--32 environment steps; Fig.~\ref{fig:trust_dynamics}(e) displays $h_s/C$, ranging from 1 to 4 action chunks.
A non-multiple of $C$ uses only the valid portion of the final chunk, so its number of full-chunk equivalents can be fractional.
When $h_s>C$, the policy and world model are recursively rolled out inside the simulator until the scheduled horizon is reached.
This design keeps imagination local when simulator reliability is low, while allowing longer multi-chunk rollouts only after the simulator becomes more reliable.

\subsection{Chunk-Level Replay Priority and Optimization Weight}
\label{app:chunk_replay_details}

At training stage $s$, \MERL{} generates a candidate imagined batch $\widetilde{\mathcal{B}}^{\mathrm{imag}}_s$ by sampling anchor observations from recent grounded trajectories and rolling out the current policy inside the evolving simulator.
Stage-level trust controls the overall imagined ratio $\rho_s$ and rollout horizon $h_s$, but chunks within $\widetilde{\mathcal{B}}^{\mathrm{imag}}_s$ can still differ in local reliability.
Chunk-level trust therefore controls two operations before policy optimization: which imagined chunks are sampled into $\mathcal{B}^{\mathrm{imag}}_s$, and how strongly each sampled chunk affects the imagined policy-gradient loss.

\paragraph{Chunk-level reliability error.}
Each candidate's local error combines visual and proxy residuals:
\begin{equation}
\varepsilon_{s,i}
=
\alpha_o\mathcal{E}^{\mathrm{obs}}_{s,i}
+
\alpha_q\mathcal{E}^{\mathrm{proxy}}_{s,i},
\label{eq:app_chunk_error}
\end{equation}
where $\mathcal{E}^{\mathrm{obs}}_{s,i}$ and $\mathcal{E}^{\mathrm{proxy}}_{s,i}$ denote estimated local visual mismatch and estimated local progress mismatch, respectively.
For exact grounded--predicted pairs, we measure:
\begin{equation}
\mathcal{E}^{\mathrm{obs}}_{s,i}
=
\sum_{j=1}^{c_i}
\omega_j d(\hat{\mathbf{o}}_{i,j},\mathbf{o}_{i,j}),
\qquad
\mathcal{E}^{\mathrm{proxy}}_{s,i}
=
\sum_{j=0}^{c_i-1}
\nu_j
\left|
\hat{q}_{i,j}-q^{\star}_{i,j}
\right|.
\label{eq:app_chunk_error_instantiation}
\end{equation}
Here $\hat{\mathbf{o}}_{i,j}$ and $\hat{q}_{i,j}$ are the imagined observation and proxy score at offset $j$ of chunk $i$, while $\mathbf{o}_{i,j}$ and $q^{\star}_{i,j}$ denote the corresponding grounded observation and success-to-go target when available.
Here $c_i\le C$ is the number of valid transitions in chunk $i$, so local residuals do not sum over the full multi-chunk horizon $h_s$.
Exact pairing requires the same anchor, instruction, and recorded action continuation; proximity of two windows alone does not constitute a pair.

\paragraph{Residual predictor for unpaired recursive chunks.}
Let $\mathbf{v}^{\mathrm{anchor}}_i$ and $\hat{\mathbf{v}}_i$ denote anchor and imagined visual latents, $d_i$ the rollout depth in action chunks, and $\mathbf{c}_s$ the stage context.
For a chunk without an exact grounded continuation, the deployed predictor computes
\begin{equation}
(\widehat{\mathcal{E}}^{\mathrm{obs}}_{s,i},\widehat{\mathcal{E}}^{\mathrm{proxy}}_{s,i})=g_\xi(\mathbf{v}^{\mathrm{anchor}}_i,\hat{\mathbf{v}}_i,\mathbf{u}_i,d_i,\mathbf{c}_s).
\label{eq:app_no_oracle_predictor}
\end{equation}
After each update of $(\mathcal{W}_\phi,\mathcal{Q}_\psi)$, we replay exact stored grounded--predicted windows to obtain residual targets from Eq.~\eqref{eq:app_chunk_error_instantiation}, refresh $g_\xi$, and freeze it for imagination and policy optimization.
For recursive calibration windows, recorded actions can be unrolled from the anchor through predicted intermediate observations, while stored futures supply the residual labels.
At deployment, the current policy instead samples later chunks from predicted observations; those unpaired chunks use predicted residuals.
Both residual outputs enter $\widehat{\varepsilon}_{s,i}=\alpha_o\widehat{\mathcal{E}}^{\mathrm{obs}}_{s,i}+\alpha_q\widehat{\mathcal{E}}^{\mathrm{proxy}}_{s,i}$ and the same replay/weighting rules below.
Future grounded observations are never predictor inputs or queried to score these recursive chunks.
The depth validation in App.~\ref{app:no_oracle_validation} holds matched futures out of scoring and uses them only afterward to measure prediction quality.
The Residual predictor in Tab.~\ref{tab:app_trust_estimator_validity} is this no-oracle component; Combined \MERL{} uses visual/proxy fusion with measured residuals where exact pairs exist and predicted residuals otherwise.
It is not a separate source of oracle information.
A smaller estimated error indicates higher predicted reliability, subject to calibration coverage.

\paragraph{Replay probability.}
Each chunk receives the inverse-error priority
\begin{equation}
a_{s,i}
=
(\varepsilon_{s,i}+\zeta_p)^{-\alpha_p},
\label{eq:app_priority_score}
\end{equation}
where $\zeta_p>0$ prevents numerical singularity and $\alpha_p\ge0$ controls the strength of reliability prioritization.
Normalizing these candidate priorities gives the replay probability used for chunk sampling:
\begin{equation}
p_{s,i}
=
\frac{a_{s,i}}
{\sum_{m\in\widetilde{\mathcal{B}}^{\mathrm{imag}}_s}a_{s,m}}.
\label{eq:app_replay_priority}
\end{equation}
Thus, $p_{s,i}$ is a sampling probability over candidate imagined chunks, not a rank.
When $\alpha_p=0$, imagined chunks are sampled uniformly; larger $\alpha_p$ concentrates replay on lower-error chunks.

\paragraph{Optimization weight.}
The per-chunk policy weight depends on reliability:
\begin{equation}
w_{s,i}
=
\mathrm{clip}
\left(
\exp(-\eta_w\varepsilon_{s,i}),
w_{\min},
1
\right),
\label{eq:app_trust_weight}
\end{equation}
where $\eta_w\ge0$ controls the decay rate and $w_{\min}\in(0,1]$ prevents the imagined branch from vanishing entirely.
The replay probability $p_{s,i}$ controls whether an imagined chunk is sampled into $\mathcal{B}^{\mathrm{imag}}_s$, whereas the optimization weight $w_{s,i}$ controls how strongly the sampled chunk contributes to the imagined loss.
This separation lets \MERL{} bias replay toward reliable imagination while still reducing gradient influence from uncertain chunks.

\subsection{Full Policy Objective}
\label{app:policy_objective_details}

At training stage $s$, \MERL{} constructs a grounded batch $\mathcal{B}^{\mathrm{real}}_s$ and an imagined batch $\mathcal{B}^{\mathrm{imag}}_s$ for policy optimization.
The grounded batch is sampled from grounded interaction data, while the imagined batch is sampled from the candidate imagined batch $\widetilde{\mathcal{B}}^{\mathrm{imag}}_s$ according to the replay probabilities $p_{s,i}$.
The two batches are sampled independently.
Grounded--imagined pairing is used only for reliability calibration when paired continuations are available; it is not required for policy optimization.

\paragraph{Mixed objective.}
We combine grounded and imagined policy updates as
\begin{equation}
\mathcal{L}^{(s)}_{\mathrm{MERL}}
=
(1-\rho_s)\mathcal{L}^{(s)}_{\mathrm{real}}
+
\rho_s\mathcal{L}^{(s)}_{\mathrm{imag}}.
\label{eq:app_merl_policy_obj}
\end{equation}
Here $\rho_s$ sets the stage-level mixture of grounded and imagined experience; $p_{s,i}$ and $w_{s,i}$ respectively control local replay admission and gradient influence.

\paragraph{Policy ratio.}
For chunk $i$ and valid action token $\ell$, the current-to-old policy probability ratio is
\begin{equation}
r_{i,\ell}(\theta)
=
\frac{
\pi_{\theta}(a_{i,\ell}\mid\mathbf{o}_i,\mathbf{x}_i,\mathbf{a}_{i,<\ell})
}{
\pi_{\theta_{\mathrm{old}}}(a_{i,\ell}\mid\mathbf{o}_i,\mathbf{x}_i,\mathbf{a}_{i,<\ell})
}.
\label{eq:app_policy_ratio}
\end{equation}
The index $\ell$ denotes a valid categorical action token; $\mathbf{o}_i$ is the grounded or predicted conditioning observation, and $\mathbf{a}_{i,<\ell}$ is prior-token context when used by the action head.
Padding and invalid action elements are excluded from the token sum.
Decoded/denormalized continuous commands are used for execution, and their densities are not substituted for categorical token probabilities.
We write the per-chunk clipped token objective in the following compact form:
\begin{equation}
\ell_{\mathrm{clip}}(i;\theta,\hat{A}_i)
=
-
\sum_{\ell}
\min
\left(
r_{i,\ell}(\theta)\hat{A}_i,
\mathrm{clip}
\left(
r_{i,\ell}(\theta),
1-\epsilon_{\mathrm{clip}},
1+\epsilon_{\mathrm{clip}}
\right)
\hat{A}_i
\right),
\label{eq:app_clipped_chunk_loss}
\end{equation}
where $\hat{A}_i$ is the group-relative advantage associated with chunk $i$, and $\epsilon_{\mathrm{clip}}$ is the clipped policy-gradient threshold.
The clipping symbol $\epsilon_{\mathrm{clip}}$ is distinct from the visual discrepancy $\delta_{s,t,j}$.

\paragraph{Grounded branch.}
The grounded branch averages the clipped objective over grounded chunks:
\begin{equation}
\mathcal{L}^{(s)}_{\mathrm{real}}
=
\frac{1}{|\mathcal{B}^{\mathrm{real}}_s|}
\sum_{i\in\mathcal{B}^{\mathrm{real}}_s}
\ell_{\mathrm{clip}}(i;\theta,\hat{A}^{\mathrm{real}}_i).
\label{eq:app_real_obj}
\end{equation}
Grounded advantages $\hat{A}^{\mathrm{real}}_i$ use the same environment feedback or reward interface as MFRL.

\paragraph{Imagined branch.}
We weight the shared clipped objective by local trust:
\begin{equation}
\mathcal{L}^{(s)}_{\mathrm{imag}}
=
\frac{1}{|\mathcal{B}^{\mathrm{imag}}_s|}
\sum_{i\in\mathcal{B}^{\mathrm{imag}}_s}
w_{s,i}\,
\ell_{\mathrm{clip}}(i;\theta,\hat{A}^{\mathrm{imag}}_i).
\label{eq:app_imag_obj}
\end{equation}
The imagined advantage $\hat{A}^{\mathrm{imag}}_i$ is computed from simulator-generated progress signals.
Because $\mathcal{B}^{\mathrm{imag}}_s$ is already sampled from $\widetilde{\mathcal{B}}^{\mathrm{imag}}_s$ according to $p_{s,i}$, Eq.~\eqref{eq:app_imag_obj} should be read as a trust-weighted imagined update rather than an unbiased estimator of uniform imagined replay.
This is intentional: \MERL{} uses chunk-level trust to bias both replay admission and gradient contribution toward more reliable imagined experience.
The GRPO-style instantiation shares its optimizer with the controls; trust acts through replay admission and loss weighting.

\subsection{Full Training Algorithm}
\label{app:full_algorithm}

Algorithm~\ref{alg:merl_full} summarizes the complete online training procedure of \MERL{}.
To keep notation consistent with the main text, the algorithm uses $\mathcal{B}^{\mathrm{real}}_s$ for grounded batches, $\widetilde{\mathcal{B}}^{\mathrm{imag}}_s$ for candidate imagined chunks, and $\mathcal{B}^{\mathrm{imag}}_s$ for imagined chunks sampled into policy optimization.

\begin{algorithm}[t]
\caption{Full MERL Online Training Procedure}
\label{alg:merl_full}
\begin{algorithmic}[1]
\REQUIRE Pretrained policy $\pi_\theta$, evolving simulator $(\mathcal{W}_\phi,\mathcal{Q}_\psi)$, perturbed environment $\mathcal{E}$
\REQUIRE Scheduler hyperparameters $\varepsilon_{\mathrm{ref}},\kappa,\rho_{\min},\rho_{\max},h_{\min},h_{\max},H_{\max},\beta_\varepsilon$
\REQUIRE Replay hyperparameters $\alpha_p,\eta_w,\zeta_p,w_{\min}$, temporal weights $\{\omega_j\},\{\nu_j\}$, clipping threshold $\epsilon_{\mathrm{clip}}$, total training stages $S$
\STATE Initialize smoothed stage-level error $\tilde{\varepsilon}_0$
\FOR{$s=1,2,\ldots,S$}
    \STATE Collect grounded trajectories using current policy $\pi_\theta$ in environment $\mathcal{E}$
    \STATE Construct grounded batch $\mathcal{B}^{\mathrm{real}}_s$
    \STATE Update world model $\mathcal{W}_\phi$ with $\mathcal{L}^{(s)}_{\mathrm{wm}}$ in Eq.~\eqref{eq:app_wm_loss}
    \STATE Update reward proxy head $\mathcal{Q}_\psi$ with $\mathcal{L}^{(s)}_{\mathrm{proxy}}$ in Eq.~\eqref{eq:app_proxy_loss}
    \STATE Replay stored exact grounded--predicted windows; compute $\bar{\varepsilon}_s$ with Eq.~\eqref{eq:app_stage_error}
    \STATE Refresh $g_\xi$ on measured residuals; freeze simulator and predictor for imagination/policy update
    \STATE Smooth it to obtain $\tilde{\varepsilon}_s$ using Eq.~\eqref{eq:app_error_smoothing}
    \STATE Compute confidence score $\chi_s$ using Eq.~\eqref{eq:app_confidence_score}
    \STATE Set imagined ratio $\rho_s$ and rollout horizon $h_s$ using Eqs.~\eqref{eq:app_ratio_schedule}--\eqref{eq:app_horizon_schedule}
    \STATE Generate candidate imagined batch $\widetilde{\mathcal{B}}^{\mathrm{imag}}_s$ by rolling out $\pi_\theta$ inside $(\mathcal{W}_\phi,\mathcal{Q}_\psi)$ for horizon $h_s$
    \FOR{each imagined chunk $i\in\widetilde{\mathcal{B}}^{\mathrm{imag}}_s$}
        \STATE Estimate chunk-level reliability error $\varepsilon_{s,i}$ using Eq.~\eqref{eq:app_chunk_error}
        \STATE Compute unnormalized priority score $a_{s,i}$ using Eq.~\eqref{eq:app_priority_score}
        \STATE Compute optimization weight $w_{s,i}$ using Eq.~\eqref{eq:app_trust_weight}
    \ENDFOR
    \STATE Normalize all candidate scores to $p_{s,i}$ using Eq.~\eqref{eq:app_replay_priority}
    \STATE Sample imagined batch $\mathcal{B}^{\mathrm{imag}}_s$ from $\widetilde{\mathcal{B}}^{\mathrm{imag}}_s$ according to $p_{s,i}$
    \STATE Update policy $\pi_\theta$ using $\mathcal{B}^{\mathrm{real}}_s$, $\mathcal{B}^{\mathrm{imag}}_s$, and the mixed objective in Eq.~\eqref{eq:app_merl_policy_obj}
\ENDFOR
\RETURN Refined policy $\pi_\theta$
\end{algorithmic}
\end{algorithm}

\section{Theoretical Proofs}
\label{app:proofs}

This appendix provides the proofs and auxiliary bias arguments used to justify the trust mechanisms in \MERL{}.
The propositions establish conditional bias bounds rather than convergence guarantees.
They show how visual prediction discrepancies propagate into proxy supervision and how this bias depends on imagined-data usage and rollout horizon.

\subsection{Proof of Proposition 1}
\label{app:proof_prop1}

Proposition~1 explains why the imagined rollout horizon should depend on simulator reliability.
The reward proxy head is supervised on grounded observation-action pairs, but during imagination it is evaluated on simulator-predicted observations.
The proposition bounds the resulting proxy-supervision bias by visual prediction discrepancies.

At anchor $t$ and horizon $h$, the discounted aggregate on imagined observations is
\begin{equation}
\hat{q}_{t}^{(h)}
:=
\sum_{j=1}^{h}
\gamma_q^{\,j-1}
\mathcal{Q}_{\psi}(\hat{\mathbf{o}}_{t+j},\mathbf{u}_{t+j}),
\label{eq:app_imagined_proxy_aggregate}
\end{equation}
and define the corresponding analytical aggregate on grounded observations as
\begin{equation}
q_{t}^{(h)}
:=
\sum_{j=1}^{h}
\gamma_q^{\,j-1}
\mathcal{Q}_{\psi}(\mathbf{o}_{t+j},\mathbf{u}_{t+j}).
\label{eq:app_grounded_proxy_aggregate}
\end{equation}
Both aggregates use the same anchor time, horizon, discount factor, and action continuation.
They differ only in whether the proxy head is evaluated on simulator-predicted observations or grounded observations.
The grounded aggregate serves analysis, not online training.

Assume that $\mathcal{Q}_{\psi}$ is $L_q$-Lipschitz with respect to observation features when the action input is fixed.
For any two observations $\mathbf{a},\mathbf{b}$ and the same action input $\mathbf{u}$,
\begin{equation}
\left|
\mathcal{Q}_{\psi}(\mathbf{a},\mathbf{u})
-
\mathcal{Q}_{\psi}(\mathbf{b},\mathbf{u})
\right|
\le
L_q \Delta(\mathbf{a},\mathbf{b}),
\label{eq:app_lipschitz_proxy}
\end{equation}
Here $\Delta(\mathbf{a},\mathbf{b})$ measures discrepancy in observation-feature space.
At stage $s$ and anchor $t$, define the pointwise visual prediction error at relative offset $j$ as
\begin{equation}
\delta_{s,t,j}
:=
\Delta(\hat{\mathbf{o}}_{t+j},\mathbf{o}_{t+j})
\label{eq:app_local_prediction_discrepancy}
\end{equation}
\begin{align}
\left|
\hat{q}_{t}^{(h)} - q_{t}^{(h)}
\right|
&=
\left|
\sum_{j=1}^{h}
\gamma_q^{\,j-1}
\Big(
\mathcal{Q}_{\psi}(\hat{\mathbf{o}}_{t+j},\mathbf{u}_{t+j})
-
\mathcal{Q}_{\psi}(\mathbf{o}_{t+j},\mathbf{u}_{t+j})
\Big)
\right|
\\
&\le
\sum_{j=1}^{h}
\gamma_q^{\,j-1}
\left|
\mathcal{Q}_{\psi}(\hat{\mathbf{o}}_{t+j},\mathbf{u}_{t+j})
-
\mathcal{Q}_{\psi}(\mathbf{o}_{t+j},\mathbf{u}_{t+j})
\right|
\\
&\le
L_q
\sum_{j=1}^{h}
\gamma_q^{\,j-1}
\Delta(\hat{\mathbf{o}}_{t+j},\mathbf{o}_{t+j})
\\
&=
L_q
\sum_{j=1}^{h}
\gamma_q^{\,j-1}
\delta_{s,t,j}.
\end{align}
The first inequality uses the triangle inequality; the second uses the Lipschitz condition in Eq.~\eqref{eq:app_lipschitz_proxy}.
Together, these inequalities give the discounted proxy-supervision bias bound:
\begin{equation}
\left|
\hat{q}_{t}^{(h)} - q_{t}^{(h)}
\right|
\le
L_q
\sum_{j=1}^{h}
\gamma_q^{\,j-1}\delta_{s,t,j}.
\end{equation}
\hfill $\square$

\paragraph{Interpretation.}
The bound shows that imagined proxy bias is controlled by a discounted accumulation of visual prediction discrepancies.
Even when the proxy head is well supervised on grounded data, evaluating it on inaccurate imagined observations can distort the progress signal.
This accumulation motivates reliability-dependent scheduling of $h_s$.

\subsection{Proof of Proposition 2}
\label{app:proof_prop2}

Proposition~2 explains why stage-level trust controls both the imagined-data ratio $\rho_s$ and the imagined rollout horizon $h_s$.
The statement should be interpreted as a scaling argument rather than a convergence theorem: if the discrepancy of imagined supervision grows with both how much imagination is used and how far it is rolled out, then scheduling only one of the two is insufficient.

Assume the stage-level imagined-supervision discrepancy admits the following bound:
\begin{equation}
\Delta_s
\le
K_{\Delta}\,\rho_s
\sum_{j=1}^{h_s}\delta_{s,j},
\label{eq:app_prop2_assumption}
\end{equation}
where $K_{\Delta}>0$ is a task-dependent constant, $\rho_s$ is the imagined-data ratio, $h_s$ is the imagined rollout horizon, and $\delta_{s,j}$ is the stage-level visual prediction discrepancy at relative offset $j$.
The quantity $\delta_{s,j}$ can be understood as an aggregate or upper bound of $\delta_{s,t,j}$ over sampled anchors.

Assume the smoothed reliability error bounds every discrepancy within the horizon:
\begin{equation}
\delta_{s,j}\le \tilde{\varepsilon}_s,
\qquad
\forall j \le h_s.
\label{eq:app_prop2_uniform_upper_bound}
\end{equation}
Here $\tilde{\varepsilon}_s$ is the normalized and smoothed stage-level reliability error used by the scheduler.
Substitution of Eq.~\eqref{eq:app_prop2_uniform_upper_bound} into Eq.~\eqref{eq:app_prop2_assumption} bounds the aggregate discrepancy by
\begin{align}
\Delta_s
&\le
K_{\Delta}\,\rho_s
\sum_{j=1}^{h_s}\tilde{\varepsilon}_s
\\
&=
K_{\Delta}\,\rho_s\,h_s\,\tilde{\varepsilon}_s.
\end{align}
Consequently, the aggregate discrepancy satisfies the stated joint scheduling bound:
\begin{equation}
\Delta_s
\le
K_{\Delta}\,\rho_s\,h_s\,\tilde{\varepsilon}_s,
\end{equation}
This establishes the joint ratio/horizon bound stated in Proposition~2. \hfill $\square$

\paragraph{Interpretation.}
The bound makes explicit that imagined-supervision discrepancy scales with three factors: simulator reliability error $\tilde{\varepsilon}_s$, imagined-data ratio $\rho_s$, and rollout horizon $h_s$.
Reducing only $\rho_s$ can still allow long biased rollouts, while reducing only $h_s$ can still admit too much unreliable imagined data.
\MERL{} therefore couples the ratio and horizon through a shared reliability signal.

\subsection{Reliability-Aware Replay Bias Bound}
\label{app:proof_prop3}

The following proposition complements the stage-level analysis by justifying the chunk-level replay and weighting rule used in \MERL{}.
It concerns local heterogeneity among generated imagined chunks and is omitted from the main paper for space.

\textbf{Proposition 3.}
Assume the bias of chunk $i$ at stage $s$ admits the following local reliability bound:
\begin{equation}
b_{s,i}
\le
K_b\varepsilon_{s,i},
\label{eq:app_chunk_bias_assumption}
\end{equation}
for some constant $K_b>0$, where $\varepsilon_{s,i}$ is the estimated chunk-level reliability error.
Let
\begin{equation}
\bar{w}_{s,i}
=
\frac{w_{s,i}}{\sum_m w_{s,m}}
\label{eq:app_normalized_weight}
\end{equation}
denote normalized weights for a per-unit-weight bias comparison.
Here $B^{\mathrm{imag}}_s$ refers to the bias of this normalized weighted mixture; the policy loss retains its chunk-count normalization in Eq.~\eqref{eq:app_imag_obj}.
Under the stated chunk-bias assumption, the aggregate bias satisfies
\begin{equation}
B^{\mathrm{imag}}_s
\le
K_b
\sum_i
\bar{w}_{s,i}\varepsilon_{s,i}.
\label{eq:app_weighted_bias_bound}
\end{equation}
For the comparison with uniform replay, suppose the chunk errors are ordered as
\begin{equation}
\varepsilon_{s,(1)}
\le
\varepsilon_{s,(2)}
\le
\cdots
\le
\varepsilon_{s,(N)}
\label{eq:app_sorted_errors}
\end{equation}
and the corresponding normalized weights are sorted in the opposite order,
\begin{equation}
\bar{w}_{s,(1)}
\ge
\bar{w}_{s,(2)}
\ge
\cdots
\ge
\bar{w}_{s,(N)},
\qquad
\sum_{i=1}^{N}\bar{w}_{s,(i)}=1,
\label{eq:app_sorted_weights}
\end{equation}
The reliability-weighted error then cannot exceed its uniform-weight counterpart:
\begin{equation}
\sum_{i=1}^{N}
\bar{w}_{s,(i)}
\varepsilon_{s,(i)}
\le
\frac{1}{N}
\sum_{i=1}^{N}
\varepsilon_{s,(i)}.
\label{eq:app_replay_no_worse}
\end{equation}
For monotone, well-calibrated weights, non-uniform chunk errors yield an aggregate bias bound at least as tight as that under uniform weighting, and typically tighter.

\paragraph{Proof.}
We define the imagined-branch bias as the normalized weighted sum of chunk biases:
\begin{equation}
B^{\mathrm{imag}}_s
:=
\sum_i
\bar{w}_{s,i} b_{s,i}.
\end{equation}
Using the chunk-wise bias assumption in Eq.~\eqref{eq:app_chunk_bias_assumption}, we obtain
\begin{align}
B^{\mathrm{imag}}_s
&=
\sum_i
\bar{w}_{s,i} b_{s,i}
\\
&\le
\sum_i
\bar{w}_{s,i} K_b\varepsilon_{s,i}
\\
&=
K_b
\sum_i
\bar{w}_{s,i}\varepsilon_{s,i},
\end{align}
This establishes the reliability-weighted imagined-branch bias bound in Eq.~\eqref{eq:app_weighted_bias_bound}.

To compare with uniform weighting, let $u_i:=1/N$.
Consider the sorted errors $\varepsilon_{s,(1)}\le\cdots\le\varepsilon_{s,(N)}$ and oppositely sorted normalized weights $\bar{w}_{s,(1)}\ge\cdots\ge\bar{w}_{s,(N)}$.
Define the cumulative deviation
\begin{equation}
S_k
:=
\sum_{i=1}^{k}
\left(
\bar{w}_{s,(i)}
-
\frac{1}{N}
\right),
\qquad
k=1,\ldots,N-1.
\end{equation}
Because $\{\bar{w}_{s,(i)}\}$ is sorted in decreasing order and sums to one, it majorizes the uniform weights, which implies $S_k\ge0$ for all $k$.
By Abel summation,
\begin{equation}
\sum_{i=1}^{N}
\left(
\bar{w}_{s,(i)}
-
\frac{1}{N}
\right)
\varepsilon_{s,(i)}
=
\sum_{k=1}^{N-1}
S_k
\left(
\varepsilon_{s,(k)}-\varepsilon_{s,(k+1)}
\right).
\end{equation}
Since $S_k\ge0$ and $\varepsilon_{s,(k)}-\varepsilon_{s,(k+1)}\le0$, each term on the right-hand side is non-positive.
Therefore,
\begin{equation}
\sum_{i=1}^{N}
\bar{w}_{s,(i)}\varepsilon_{s,(i)}
\le
\frac{1}{N}
\sum_{i=1}^{N}
\varepsilon_{s,(i)}.
\end{equation}
This proves the comparison of reliability-aware and uniform weighting in Eq.~\eqref{eq:app_replay_no_worse}. \hfill $\square$

\paragraph{Interpretation.}
Proposition~3 motivates chunk-level trust: generated chunks can have heterogeneous local errors even when average simulator reliability is acceptable.
If the trust weights are reasonably calibrated, assigning larger optimization weight to lower-error chunks yields a tighter aggregate bias upper bound than uniform weighting.
This supports the joint use of stage-level scheduling and chunk-level replay and weighting for reliable policy refinement in \MERL{}.

\section{Experimental Details}
\label{app:exp_details}

This appendix provides implementation details for the experiments in the main paper.
We describe the benchmark scope, policy backbone, evaluation metrics, training infrastructure, and baseline implementations.
The main text reports compact summaries, while this appendix specifies how the metrics are computed and how the online training budget is accounted for.

\subsection{Benchmark and Perturbation Scope}
\label{app:benchmark_scope}

We evaluate \MERL{} on LIBERO-PRO~\cite{zhou2025libero}, which extends LIBERO-style language-conditioned manipulation tasks with controlled deployment-time perturbations.
The main experiments cover four perturbed suites: Spatial, Object, Goal, and Long.
These suites keep task semantics comparable while shifting spatial layouts, object states, goal conditions, or long-horizon execution contexts.
This protocol evaluates online VLA post-training under deployment distribution shifts rather than open-ended acquisition of entirely new task categories.
It directly tests the failure mode targeted by \MERL{}: frozen simulators become stale under policy-induced shifts, while online-updated simulators still require reliability auditing before policy optimization.

\subsection{Policy Backbone and Action Representation}
\label{app:policy_backbone_details}

All methods use the same OpenVLA-OFT~\cite{kim2025fine} policy checkpoint as initialization.
In our refinement implementation, the policy uses a tokenized action head compatible with clipped policy-gradient optimization.
In particular, \MERL{} optimizes a GRPO-style objective that requires old-policy and new-policy action probabilities to compute policy ratios.
Token-based VLA policies naturally expose such action distributions and support stochastic rollout sampling.
By contrast, flow-matching or diffusion-action VLAs require additional likelihood estimation, trajectory-level reparameterization, or a different RL objective to compute comparable policy ratios.
Extending \MERL{} to those continuous-action generative policies is conceptually possible, but it is orthogonal to the trust-calibrated imagination mechanism and left for future work.

Following the OpenVLA-OFT setting, the policy predicts an action chunk of length $C=8$ with a 7-dimensional action interface.
We use a single global RGB observation and the language instruction as policy input in simulation.
Unless otherwise specified, proprioceptive inputs are disabled to keep the setup aligned with the global-camera deployment setting used in our real-robot validation.

\subsection{Evaluation Metrics}
\label{app:metric_definitions}

Let $\mathcal{G}=\{\mathrm{Spatial},\mathrm{Object},\mathrm{Goal},\mathrm{Long}\}$ denote the set of perturbed evaluation suites.
For suite $g\in\mathcal{G}$, let $\mathcal{T}_g$ denote its task set.
We evaluate a policy at checkpoints indexed by $k=0,\ldots,K$.
Let $n_k$ be the cumulative outer-loop refinement stage when checkpoint $k$ is evaluated; it does not count environment transitions, optimizer minibatches, or task-horizon steps.
All online methods are evaluated at the same checkpoint set $\{n_k\}_{k=0}^{K}$ under the same online training budget.
For model-based methods, simulator-generated rollouts are part of the training procedure; S2T therefore measures learning speed along the common online training axis rather than claiming a pure real-interaction count.

\paragraph{Success rate.}
For task $m\in\mathcal{T}_g$ and evaluation rollout $r$, let
$y_{g,m,r}^{(k)}\in\{0,1\}$ indicate whether checkpoint $k$ succeeds.
The success rate of suite $g$ at checkpoint $k$ is
\begin{equation}
\mathrm{SR}_g(k)
=
100\cdot
\frac{1}{|\mathcal{T}_g|R}
\sum_{m\in\mathcal{T}_g}
\sum_{r=1}^{R}
y_{g,m,r}^{(k)} ,
\label{eq:app_sr}
\end{equation}
where $R$ is the number of evaluation rollouts per task.
Final average SR across the four suites is
\begin{equation}
\mathrm{Avg.\ SR}
=
\frac{1}{|\mathcal{G}|}
\sum_{g\in\mathcal{G}}
\mathrm{SR}_g(K).
\label{eq:app_avg_sr}
\end{equation}

\paragraph{Area under the training curve.}
AUC measures online learning efficiency over the full refinement trajectory.
For suite $g$, we integrate the SR curve by the normalized trapezoidal rule:
\begin{equation}
\mathrm{AUC}_g
=
\frac{1}{n_K-n_0}
\sum_{k=1}^{K}
\frac{\mathrm{SR}_g(k-1)+\mathrm{SR}_g(k)}{2}
\left(n_k-n_{k-1}\right).
\label{eq:app_auc}
\end{equation}
We then average the suite-wise AUC values to obtain the overall learning-curve metric:
\begin{equation}
\mathrm{AUC}
=
\frac{1}{|\mathcal{G}|}
\sum_{g\in\mathcal{G}}
\mathrm{AUC}_g.
\label{eq:app_avg_auc}
\end{equation}
Since SR is expressed in percentage, AUC is also reported on a percentage scale.
This metric rewards methods that improve early and remain strong throughout online refinement.

\paragraph{Steps to threshold.}
S2T measures how many online training steps are required to reach a target success level.
For suite $g$ and threshold $\eta\in[0,100]$, define
\begin{equation}
\mathrm{S2T}_g(\eta)
=
\inf
\left\{
n \;|\; \mathrm{SR}_g(n)\ge \eta
\right\},
\label{eq:app_s2t}
\end{equation}
where $\mathrm{SR}_g(n)$ denotes the success-rate curve as a function of cumulative online training steps.
In practice, policies are evaluated only at discrete checkpoints $\{n_k\}_{k=0}^{K}$.
If $\mathrm{SR}_g(k)<\eta\le \mathrm{SR}_g(k+1)$, we estimate the threshold-crossing point by linear interpolation:
\begin{equation}
\mathrm{S2T}_g(\eta)
=
n_k+
\frac{\eta-\mathrm{SR}_g(k)}
{\mathrm{SR}_g(k+1)-\mathrm{SR}_g(k)}
\left(n_{k+1}-n_k\right).
\label{eq:app_s2t_interp}
\end{equation}
If a method does not reach the threshold within the training budget, we mark it as not reached in threshold-sweep plots and use the maximum budget for aggregate summaries.
Although online training steps are evaluated at discrete checkpoints, S2T can be fractional because the threshold-crossing point is interpolated between checkpoints and later averaged across suites.
Smaller S2T therefore denotes faster threshold attainment under the shared protocol.

\paragraph{High-threshold summary.}
The main table reports a compact high-threshold summary, denoted as S2T-H.
We use $\eta_g=80$ for Spatial, Object, and Goal, and $\eta_g=65$ for Long:
\begin{equation}
\mathrm{S2T\mbox{-}H}
=
\frac{1}{|\mathcal{G}|}
\sum_{g\in\mathcal{G}}
\mathrm{S2T}_g(\eta_g),
\qquad
\eta_g=
\begin{cases}
80, & g\in\{\mathrm{Spatial},\mathrm{Object},\mathrm{Goal}\},\\
65, & g=\mathrm{Long}.
\end{cases}
\label{eq:app_s2th}
\end{equation}
The lower threshold for Long reflects its harder long-horizon structure under the same training budget.
All thresholds are fixed before method comparison and are shared across methods.
Because each suite-level S2T is interpolated and S2T-H averages across suites, the reported S2T-H values are shown with one decimal place.
We report full threshold-sweep curves to show that the comparison is not determined by a single threshold choice.

\paragraph{Metrics for SFT.}
SFT is evaluated as a fixed pretrained policy without online refinement.
Therefore, we report its final SR as a reference but do not report AUC or S2T in the main table.
Learning-curve plots show SFT as a horizontal reference.

\subsection{Grounded Budgets and Statistical Comparisons}
\label{app:budget_accounting}

The common training axis counts outer-loop refinement stages.
After stage $s$, every online method has collected $6s$ grounded trajectories subject to the same $\le3072s$ transition cap.
Trajectory lengths can differ through termination; the cap is a shared interaction allowance, not a claim that every method realizes an identical transition count.
Grounded--predicted calibration and residual-predictor refresh reuse stored windows and add zero environment interaction.
Imagined rollouts are model computation and do not consume this grounded interaction budget.
Consequently, the $10.9\%$ S2T-H reduction against matched MFRL describes faster threshold attainment under the common interaction protocol, rather than a measured wall-clock or environment-transition saving.

Tab.~\ref{tab:external_results} reports means and standard deviations from three training seeds.
SimpleVLA-RL, WMPO, WoVR, and \MERL{} share SFT initialization, split, grounded budget, action chunk, optimizer/reward interface, checkpoints, and evaluation.
WMPO and WoVR are same-protocol ports that change benchmark/model assets while retaining their defining algorithms; each WoVR seed uses an independent simulator.
The component controls in Tab.~\ref{tab:main_results} retain their original point estimates and serve a different comparison from these external algorithms.
Paired three-seed contrasts appear in Tab.~\ref{tab:paired_contrasts}.

\begin{table}[t]
\centering
\small
\setlength{\tabcolsep}{8pt}
\caption{Paired 95\% CIs over three seeds. AUC: points; S2T-H: refinement stages.}
\label{tab:paired_contrasts}
\begin{tabular}{lcr}
\toprule
Comparison & Metric & Difference [95\% CI]\\
\midrule
\MERL{} -- WoVR & $\Delta$AUC $\uparrow$ & $2.2\,[1.2,3.3]$\\
\MERL{} -- WoVR & $\Delta$S2T-H $\downarrow$ & $-9.2\,[-15.4,-3.1]$\\
\MERL{} -- matched Online-Updated MBRL & $\Delta$AUC $\uparrow$ & $5.0\,[3.8,6.2]$\\
\bottomrule
\end{tabular}
\end{table}

\subsection{Training Infrastructure}
\label{app:training_infrastructure}

All main experiments are conducted on a single node with $4\times$ NVIDIA H100 GPUs with 80GB memory.
We allocate three GPUs for actor rollout, policy inference, log-probability computation, and policy optimization, and reserve one GPU for simulator update and simulator inference.
This separation avoids resource contention between the VLA policy and the evolving simulator, and allows simulator updates to proceed without blocking all actor workers.

A standard 100-stage online run takes approximately 35 hours under our current implementation.
Each online training stage consists of grounded rollout collection, simulator update, reliability estimation, imagined rollout generation, and mixed policy optimization.
We use Ray-based distributed execution inherited from the VLA-RL training stack, with FSDP-style sharding and gradient checkpointing for the actor model.
Headless rendering uses an EGL/GLX-compatible service.

\subsection{Baseline Implementation}
\label{app:baseline_implementation}

The component controls share the same policy initialization, task suite, grounded budget, rollout temperature, action chunk length, and evaluation protocol.
The concrete external methods in Tab.~\ref{tab:external_results} are evaluated separately as same-protocol comparisons; they are not renamed component controls.

\paragraph{SFT.}
Pretrained OpenVLA-OFT is evaluated directly on perturbed LIBERO-PRO.

\paragraph{MFRL.}
MFRL performs model-free policy refinement.
It uses the same GRPO-style policy optimizer and environment reward interface as \MERL{}, but disables simulator-generated imagined rollouts.
At a matched budget, this control isolates gains from imagination.

\paragraph{Static-MBRL.}
Static-MBRL enables simulator-generated imagined rollouts but keeps the simulator frozen during online policy refinement.
It uses a fixed imagined-data ratio, a fixed imagined rollout horizon, and uniform imagined replay.
This baseline tests whether merely adding imagined data is sufficient when the simulator is not updated online.

\paragraph{Static-MBRL+Trust.}
Static-MBRL+Trust also keeps the simulator frozen, but applies the same stage-level scheduling and chunk-level replay weighting as \MERL{}.
This baseline isolates the effect of trust calibration without online simulator evolution.

\paragraph{Online-Updated MBRL.}
Online-Updated MBRL updates the simulator during policy refinement, but uses generated imagination with fixed admission and uniform replay.
This baseline corresponds to the online-updated model-based refinement paradigm discussed in the main paper and tests whether simulator update alone is sufficient.

\paragraph{\MERL{}.}
\MERL{} updates the simulator online and applies both stage-level and chunk-level trust.
Stage-level trust controls the imagined-data ratio and rollout horizon, while chunk-level trust controls replay probability and optimization weight for individual imagined chunks.

\section{Additional Ablations and Diagnostics}
\label{app:extra_ablations}

This section provides supplementary analyses omitted from the main paper due to space limits.
We focus on three aspects: simulator evolution, trust-estimator validity, and cross-suite robustness.
We do not report a separate perturbation-severity sweep because the main benchmark already evaluates heterogeneous perturbation suites under a fixed real-interaction budget.
Instead, robustness is assessed by performance consistency across Spatial, Object, Goal, and Long suites, together with the real-robot deployment validation in App.~\ref{app:real_robot_eval}.

\input{sec/component_ablation}

\subsection{Simulator Evolution Diagnostics}
\label{app:simulator_diagnostics}

A central claim of \MERL{} is that online simulator evolution improves the quality of simulator-generated imagination under the current policy-induced distribution.
We therefore evaluate the simulator separately from downstream policy performance.
The comparison is conducted on held-out grounded trajectories collected during online refinement, using the same anchor observations, instructions, and action chunks for all simulators.
This held-out comparison isolates simulator prediction quality from the effects of downstream policy optimization.

\paragraph{Quantitative metrics.}
For visual prediction, we report PSNR and SSIM as fidelity metrics, and LPIPS and FVD as perceptual and temporal consistency metrics.
For the reward proxy head, we report MAE and MSE against the truncated success-to-go target.
Higher PSNR/SSIM and lower LPIPS/FVD/MAE/MSE indicate better simulator reliability.

\begin{table}[t]
\centering
\small
\setlength{\tabcolsep}{4pt}
\caption{
Simulator diagnostics on held-out policy-induced trajectories.
The frozen simulator corresponds to Static-MBRL, while the evolving simulator is updated online during \MERL{} training.
}
\label{tab:app_simulator_diagnostics}
\begin{tabular}{lcccccc}
\toprule
Simulator & PSNR $\uparrow$ & SSIM $\uparrow$ & LPIPS $\downarrow$ & FVD $\downarrow$ & Proxy MAE $\downarrow$ & Proxy MSE $\downarrow$ \\
\midrule
Frozen simulator & 22.4 & 0.684 & 0.238 & 286.0 & 0.183 & 0.057 \\
Evolving simulator & \textbf{25.9} & \textbf{0.759} & \textbf{0.171} & \textbf{198.0} & \textbf{0.126} & \textbf{0.031} \\
\bottomrule
\end{tabular}
\end{table}

\paragraph{Results.}
Tab.~\ref{tab:app_simulator_diagnostics} shows that online simulator evolution improves both visual prediction and proxy-progress estimation.
Compared with the frozen simulator, the evolving simulator improves PSNR from $22.4$ to $25.9$, reduces LPIPS from $0.238$ to $0.171$, and reduces FVD from $286.0$ to $198.0$.
The proxy error also decreases, with MAE reduced from $0.183$ to $0.126$.
These results support the need for simulator evolution: as the policy visits states that are not fully covered by the initial data distribution, updating the simulator keeps imagined rollouts better aligned with newly induced trajectories.
However, improved average simulator quality does not imply that every imagined rollout is reliable, which motivates the trust diagnostics in App.~\ref{app:trust_estimator_diagnostics}.

\begin{figure}[t]
\centering
\includegraphics[width=0.95\linewidth]{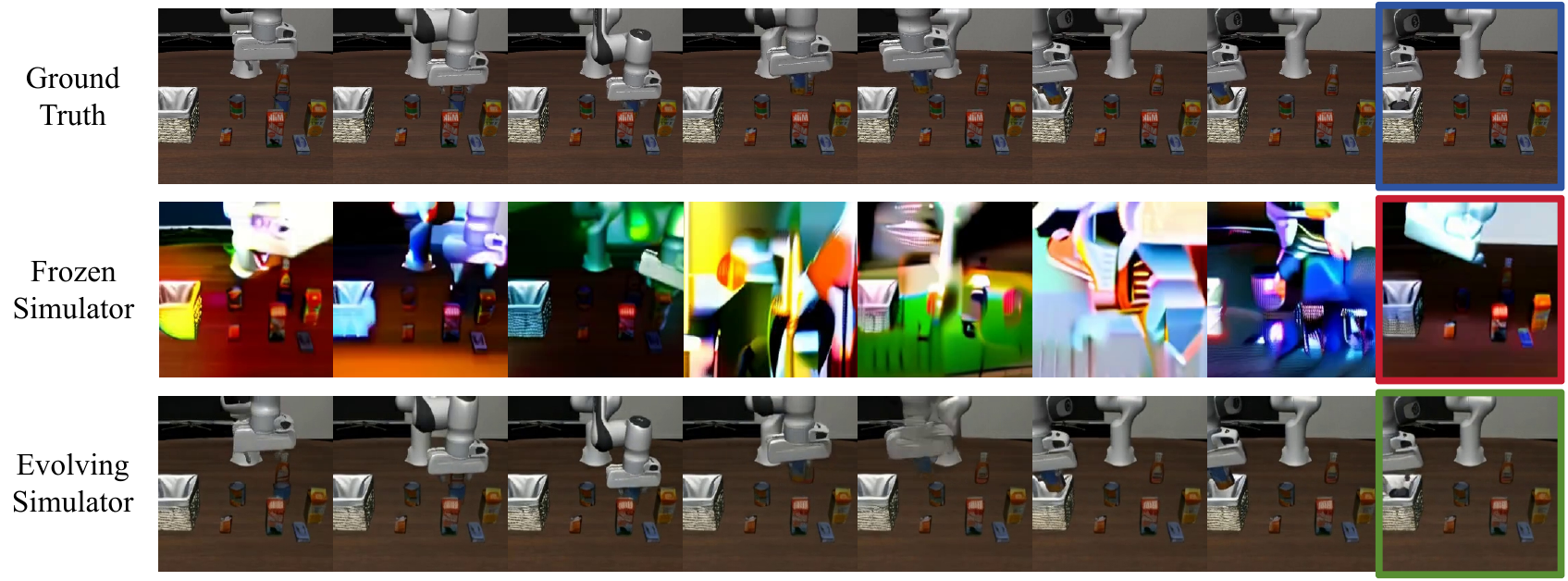}
\caption{
Qualitative comparison of simulator predictions.
Rows show ground-truth future frames, predictions from the frozen simulator, and predictions from the evolving simulator under the same rollout condition.
The evolving simulator produces visual continuations that remain closer to the ground truth over the prediction horizon.
}
\label{fig:app_wm_visualization}
\end{figure}

\paragraph{Qualitative visualization.}
Fig.~\ref{fig:app_wm_visualization} compares simulator-predicted future frames against the ground-truth trajectory under the same rollout condition.
The frozen simulator quickly departs from the real visual sequence, producing strong color artifacts, distorted robot geometry, and misplaced objects in later frames.
By contrast, the evolving simulator preserves the tabletop layout, robot pose, and object configurations more faithfully over the prediction horizon.
This visualization complements Tab.~\ref{tab:app_simulator_diagnostics} by showing that online simulator evolution improves not only aggregate prediction metrics, but also the visual reliability of imagined rollouts used for policy refinement.

\subsection{Trust-Estimator and Trust-Chain Diagnostics}
\label{app:trust_estimator_diagnostics}

We next examine whether the reliability estimates used by \MERL{} are informative.
This analysis addresses a potential concern that the trust mechanism may simply act as a heuristic schedule.
We evaluate whether estimated reliability correlates with actual rollout error, whether low-error chunks are preferentially replayed, and whether trust calibration stabilizes policy updates.

\paragraph{Estimator validity.}
We compare several reliability estimators.
For each estimator, we measure correlation with observed rollout residuals, ranking quality, calibration error, and downstream performance.
Here Rank-AUC measures how well the estimator ranks lower-error chunks ahead of higher-error chunks, ECE measures calibration error, and Top30 Low-Error Precision measures the fraction of truly low-error chunks among the top $30\%$ most trusted samples.

\begin{table}[t]
\centering
\small
\setlength{\tabcolsep}{4pt}
\caption{
Trust-estimator validity and downstream policy performance.
}
\label{tab:app_trust_estimator_validity}
\begin{tabular}{lccccc}
\toprule
Estimator & Pearson $r$ $\uparrow$ & Rank-AUC $\uparrow$ & ECE $\downarrow$ & Top30 Prec. $\uparrow$ & AUC $\uparrow$ \\
\midrule
Random / uniform & 0.02 & 0.50 & 0.176 & 0.31 & 66.1 \\
WM latent error only & 0.55 & 0.72 & 0.114 & 0.66 & 67.0 \\
Proxy residual only & 0.49 & 0.68 & 0.126 & 0.61 & 66.6 \\
Residual predictor & 0.71 & 0.82 & 0.063 & 0.79 & 69.4 \\
Combined \MERL{} estimator & \textbf{0.79} & \textbf{0.87} & \textbf{0.044} & \textbf{0.86} & \textbf{70.6} \\
\bottomrule
\end{tabular}
\end{table}

Tab.~\ref{tab:app_trust_estimator_validity} shows that the combined estimator is more predictive than single-source reliability signals.
Using only world-model latent error or proxy residual improves over random scoring, but both miss part of the local failure structure.
The combined estimator achieves the highest correlation, ranking quality, and downstream AUC, supporting both residual sources for local chunk-level trust.

\paragraph{Trust-chain diagnostics.}
We further verify whether the estimated trust signals are actually used as intended by the scheduler and replay mechanism.
Tab.~\ref{tab:app_trust_chain} reports correlations between reliability errors and the resulting schedules, replay probabilities, and policy weights.

\begin{table}[t]
\centering
\small
\setlength{\tabcolsep}{5pt}
\caption{
Trust-chain diagnostics.
Negative correlations indicate that lower estimated error leads to more imagination, longer rollout horizons, higher replay probability, or larger optimization weight.
}
\label{tab:app_trust_chain}
\begin{tabular}{lcc}
\toprule
Diagnostic link & Statistic & Value \\
\midrule
Simulator error vs. imagined ratio $\rho_s$ & Spearman & $-0.967$ \\
Simulator error vs. rollout horizon $h_s$ & Spearman & $-0.963$ \\
Chunk error vs. replay probability $p_{s,i}$ & Spearman & $-0.750$ \\
Chunk error vs. trust weight $w_{s,i}$ & Spearman & $-0.950$ \\
High-error quartile replay share / uniform share & Ratio & $0.180$ \\
KL standard deviation reduction vs. Online-Updated MBRL & Relative & $0.430$ \\
\bottomrule
\end{tabular}
\end{table}

These diagnostics confirm the intended trust chain.
At the stage level, lower simulator error increases the imagined-data ratio $\rho_s$ and permits longer rollout horizons $h_s$.
At the chunk level, lower chunk-level reliability error $\varepsilon_{s,i}$ leads to higher replay probability $p_{s,i}$ and larger trust weight $w_{s,i}$.
High-error chunks are substantially under-sampled compared with uniform replay, and policy-update variability is reduced relative to Online-Updated MBRL.
This supports the interpretation that \MERL{} improves performance by auditing imagined data before it shapes policy optimization, rather than by merely adding more imagined samples.

\input{sec/rebuttal_evidence}

\subsection{Cross-Suite Robustness}
\label{app:cross_suite_robustness}

Although we do not report a manually tuned perturbation-severity sweep, we evaluate robustness by measuring consistency across heterogeneous LIBERO-PRO perturbation suites.
This avoids introducing an additional perturbation protocol while still testing whether a method remains effective under different spatial, object, goal, and long-horizon shifts.

\begin{table}[t]
\centering
\small
\setlength{\tabcolsep}{5pt}
\caption{
Cross-suite robustness across perturbed LIBERO-PRO suites.
Avg. SR measures overall performance, Worst SR measures the weakest suite, and Std. measures sensitivity across suites.
}
\label{tab:app_cross_suite_robustness}
\begin{tabular}{lccc}
\toprule
Method & Avg. SR $\uparrow$ & Worst SR $\uparrow$ & Std. $\downarrow$ \\
\midrule
MFRL & 77.2 & 66.0 & 6.61 \\
Static-MBRL & 73.3 & 60.7 & 7.36 \\
Online-Updated MBRL & 76.2 & 66.1 & \textbf{6.08} \\
\MERL{} (Ours) & \textbf{79.7} & \textbf{68.8} & 6.37 \\
\bottomrule
\end{tabular}
\end{table}

\paragraph{Results.}
Tab.~\ref{tab:app_cross_suite_robustness} shows that \MERL{} obtains the strongest average and worst-suite success rates.
Its cross-suite standard deviation is not the lowest; this statistic reflects variation in suite difficulty and differs from across-seed uncertainty.
For robustness, the more relevant observation is that \MERL{} raises the weakest-suite performance from $66.1$ for Online-Updated MBRL and $66.0$ for MFRL to $68.8$.
This indicates that trust-calibrated imagination improves performance under the most difficult evaluated perturbation setting, rather than only improving average performance on easier suites.

\section{Real Robot Evaluation}
\label{app:real_robot_eval}

We further evaluate \MERL{} on a physical robot to examine whether trust-calibrated imagination can support real-world deployment.
This study is intended as a real-robot deployment validation rather than a full online real-world RL benchmark.
Because real-robot interaction is expensive and safety-constrained, we collect a small set of expert demonstrations, perform offline refinement on a server, and deploy the refined checkpoints back to the robot for inference-only evaluation.

\paragraph{Platform.}
Experiments are conducted on a single-arm AgileX PiPER X tabletop manipulation platform equipped with a two-finger gripper.
The robot is observed by a fixed global RGB camera, following a PiPER-family tabletop deployment setup similar to prior compositional simulation platforms~\cite{qin2026comsim}.
At deployment time, the policy receives the global camera image and the language instruction, predicts a 7-dimensional action command, and sends the command to the robot controller.
All evaluated methods use the same global RGB camera pose, workspace boundary, control interface, reset protocol, and control frequency.

\paragraph{Real-robot data collection.}
We collect expert demonstrations on two tabletop manipulation tasks:
\emph{Block-to-Box}, where the robot grasps a wooden block and places it into a box, and
\emph{Lid-to-Pot}, where the robot grasps a cooker lid and places it on the cooker.
For each task, we collect 50 demonstrations.
Each demonstration contains synchronized RGB video, executed actions, robot joint states, and gripper states.
During collection, object locations and object colors are varied to introduce moderate pose and appearance generalization while keeping the task semantics fixed.

\paragraph{Fixed-data offline refinement.}
All methods use the same 50 demonstrations per task, hardware interface, offline update budget, optimizer, and action horizon.
The demonstration dataset is fixed throughout refinement, and no additional physical policy rollouts are collected.
SFT provides the supervised policy reference; MFRL is the grounded-only refinement control on these fixed data.
Its name denotes the control family and does not imply online physical interaction in this study.
Online-Updated MBRL adapts the world model on stored robot data and uses visual imagination.
\MERL{}-Visual uses the visual-only trust configuration, with $\alpha_o=1$ and $\alpha_q=0$.
Full \MERL{} enables the reward proxy $\mathcal{Q}_\psi$ and joint visual/proxy trust.
The visual-only variant disables proxy supervision for imagined refinement, while the full variant evaluates the complete simulator/trust interface.

At each offline refinement stage, simulator adaptation and calibration reuse the demonstrations.
The policy mixes grounded demonstration chunks with imagined chunks; stage-level trust controls the amount and horizon of imagination, and chunk-level trust controls replay and loss weighting.
The residual predictor is refreshed from exact stored pairs and frozen for unpaired recursive scoring as in App.~\ref{app:chunk_replay_details}.
The server-refined checkpoints are then deployed for closed-loop evaluation.

\paragraph{Evaluation protocol.}
Each method is evaluated on 20 trials per task under the same held-out object placements and color variations.
Success requires completing the command without human intervention within the episode horizon.
Failures include incorrect approach, unstable grasp, premature release, collision, placement outside the target, or timeout.
All methods share camera pose, workspace, control interface, reset protocol, and control frequency.

\paragraph{Real-robot results.}
Tab.~\ref{tab:real_robot_success} reports all five methods, including the visual-only configuration and proxy-enabled full method.
Online-Updated MBRL achieves 13/20 Block-to-Box and 11/20 Lid-to-Pot successes ($60.0\%$ average).
\MERL{}-Visual achieves 16/20 and 15/20 ($77.5\%$), preserving the visual-only result.
Full \MERL{} achieves 17/20 and 15/20 ($80.0\%$), while SFT and MFRL obtain $35.0\%$ and $55.0\%$.
The average-SR 95\% CIs are $[22.1,50.5]$, $[39.8,69.3]$, $[44.6,73.7]$, $[62.5,87.7]$, and $[65.2,89.5]$ for SFT, MFRL, Online-Updated MBRL, \MERL{}-Visual, and full \MERL{}, respectively.
The full and visual-only intervals overlap substantially; these trials support feasibility of full-method deployment without isolating a statistically significant proxy contribution.

\begin{figure}[t]
\centering
\begin{minipage}{0.485\linewidth}
\centering\includegraphics[width=\linewidth]{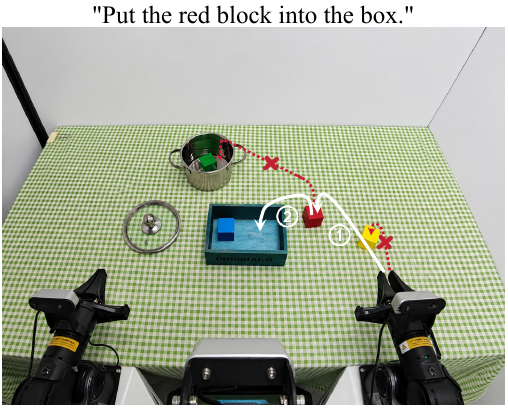}\\
{\footnotesize Block-to-Box}
\end{minipage}\hfill
\begin{minipage}{0.485\linewidth}
\centering\includegraphics[width=\linewidth]{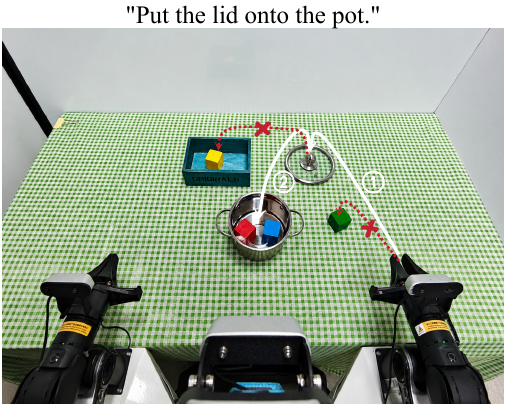}\\
{\footnotesize Lid-to-Pot}
\end{minipage}
\caption{AgileX PiPER X task setups for fixed-data adaptation and closed-loop physical evaluation.}
\label{fig:real_robot_setups}
\end{figure}

\begin{figure}[t]
\centering
\includegraphics[width=0.95\linewidth]{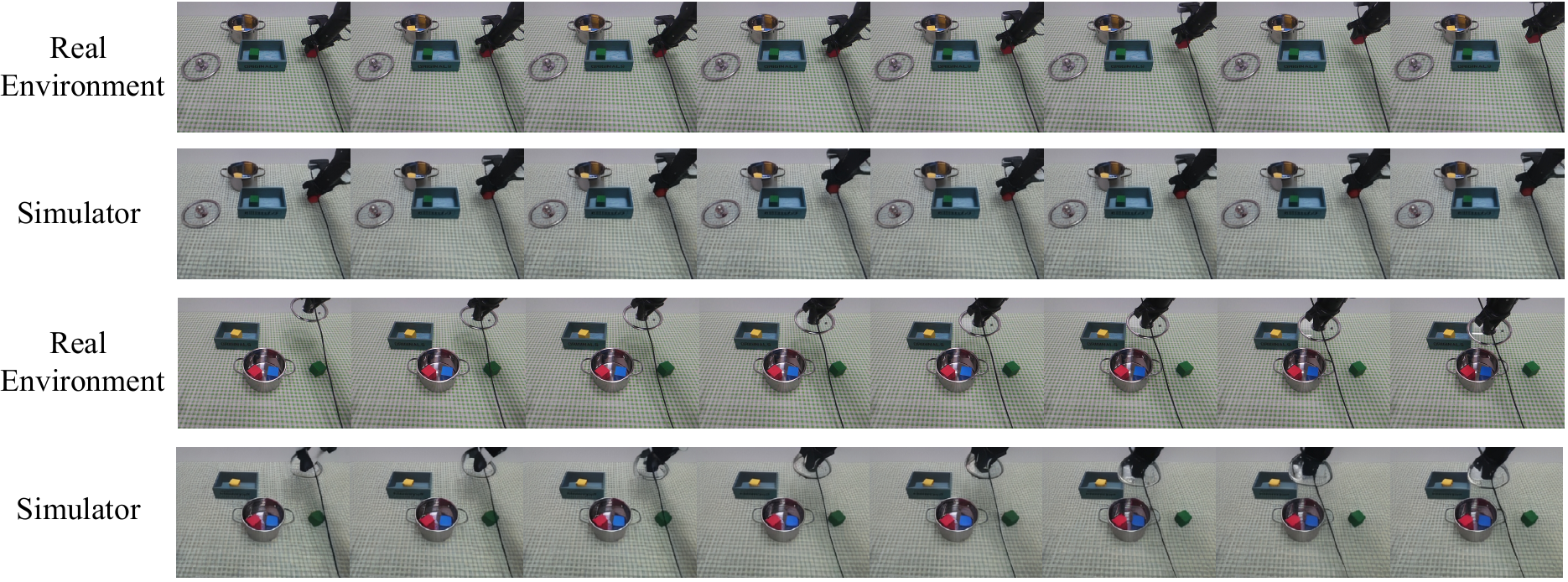}
\caption{
MERL real-robot deployment and simulator visualization on AgileX PiPER X.
From top to bottom, rows show real Block-to-Box rollout, simulator-generated Block-to-Box continuation, real Lid-to-Pot rollout, and simulator-generated Lid-to-Pot continuation.
}
\label{fig:real_robot_rollouts}
\end{figure}

\paragraph{Qualitative rollout visualization.}
Fig.~\ref{fig:real_robot_rollouts} visualizes MERL deployment on two real-robot tasks.
For each task, we show the policy's real closed-loop execution frames together with simulator-generated visual continuations conditioned on the corresponding rollout context.
The real frames demonstrate that the refined policy can approach the target, execute grasping, and complete placement under held-out object poses and appearance variations.
The simulator rows provide an aligned observation-space view of the same manipulation process, showing that the evolving simulator preserves the main scene layout, robot motion, and object-level task progress along the rollout.
This four-row visualization complements the success-rate table by showing both physical execution and the simulator imagination used by MERL during real-robot adaptation.

\paragraph{Scope and limitations.}
This study covers fixed-data offline adaptation and closed-loop physical deployment on two tabletop tasks.
It does not evaluate online physical RL or additional policy-driven data collection.
Remaining failures include fixed-camera depth ambiguity, gripper slippage, camera--robot calibration drift, and contact dynamics that short-horizon RGB prediction may not capture.
The limited trial count and overlapping intervals constrain conclusions about the proxy-specific gain.
Broader physical tasks and interactive refinement require further evaluation.

\section{Additional Related Work on World Models}
\label{app:wm_related}

This section complements the main related work by positioning the evolving simulator in \MERL{} within recent world-model paradigms.
Existing world models differ in their representation space, prediction target, and compatibility with downstream policy refinement.
The central distinction for \MERL{} is not only how a future state is predicted, but whether the predicted future can be safely used as training data for VLA policy post-training.

\paragraph{Implicit predictive world models.}
A first line of work studies implicit or latent predictive world models, where the objective is to learn predictive representations rather than reconstruct future observations in pixel space.
JEPA-style models are representative of this direction, predicting abstract latent targets while discarding unpredictable visual details.
Recent examples include V-JEPA~2, which combines large-scale self-supervised video pretraining with a small amount of robot interaction data for understanding, prediction, and planning~\cite{assran2025vjepa2}, and V-JEPA~2.1, which further improves dense visual grounding and downstream robotic planning~\cite{murlabadia2026vjepa21}.
These methods provide strong abstractions and are attractive for planning, but they are less directly aligned with our setting, where the VLA policy consumes visual observations and the refinement loop benefits from explicit generated rollouts that can be evaluated, filtered, and replayed.

\paragraph{3D spatial world models.}
A second line explicitly represents the world in 3D, using point-based, Gaussian-based, or geometry-aware structures.
This direction improves spatial consistency and physical grounding beyond 2D future prediction.
Representative examples include GWM, which uses Gaussian splatting and diffusion-based dynamics for robotic manipulation~\cite{lu2025gwm}, and PointWorld, which unifies state and action in a shared 3D point-flow space for in-the-wild robotic manipulation~\cite{huang2026pointworld}.
Such models are promising for spatial reasoning and contact-aware control, but they often require specialized reconstruction pipelines and 3D state representations.
By contrast, \MERL{} targets online VLA post-training with RGB observations and chunked actions, where a visual rollout interface is easier to integrate with existing image-conditioned policies.

\paragraph{Video-generative world models.}
\MERL{} is most closely related to video-generative world modeling.
Large-scale video generation models such as Open-Sora~2.0~\cite{peng2025opensora2} and Wan~\cite{wang2025wan} have improved the realism, controllability, and scalability of video synthesis.
Robotics and embodied-AI works further adapt video generation for action-conditioned prediction, synthetic trajectory generation, and visual world knowledge transfer, including humanoid video world models~\cite{ali2025hwm}, DreamGen-style synthetic trajectory generation~\cite{jang2025dreamgen}, and transferable visual world knowledge from large-scale videos~\cite{ren2026videoworld2}.
This branch is attractive for VLA refinement because it operates in the same observation space as image-conditioned policies and can generate rollouts that resemble downstream policy inputs.
However, video realism alone is not sufficient for policy learning.
Most video-generative world models optimize future-frame plausibility or temporal coherence, but do not by themselves decide whether a generated rollout provides reliable supervision for policy optimization.
\MERL{} augments visual imagination with a reward proxy and two-level trust that schedules, selects, and weights rollouts using simulator reliability.

\paragraph{RSSM-style world models for reinforcement learning.}
Another influential line comes from model-based reinforcement learning, starting from early neural world models~\cite{ha2018worldmodels} and recurrent state-space models such as PlaNet~\cite{hafner2019planet}, and later developed through the Dreamer family.
In this paradigm, the world model is part of the reinforcement-learning loop and usually predicts latent transitions, rewards, and continuation signals.
Representative examples include Dreamer-style latent imagination~\cite{hafner2023dreamerv3}, the Nature 2025 version of Dreamer~\cite{hafner2025dreamer}, Dreamer~4 for scalable imagination training~\cite{hafner2025dreamer4}, TWISTER for transformer-based world modeling with contrastive predictive coding~\cite{burchi2025twister}, and Dreamer-CDP for reconstruction-free deterministic representation prediction~\cite{hauri2026dreamercdp}.
These methods are natively integrated with RL because reward prediction is part of the modeling stack.
However, they typically operate in compact latent state spaces and are not designed around modern VLA policies that take high-dimensional visual observations and output chunked actions.
\MERL{} borrows the reinforcement-learning motivation of using imagination for policy improvement, while retaining a visual simulator interface compatible with VLA post-training.

\paragraph{Position of \MERL{}.}
\MERL{} is not proposed as a new world-model architecture.
Instead, it addresses a complementary question: once a visual simulator produces imagined rollouts for VLA post-training, how should those rollouts be trusted?
The evolving simulator keeps imagination aligned with newly collected grounded trajectories, while stage-level trust controls the imagined-data ratio $\rho_s$ and rollout horizon $h_s$, and chunk-level trust controls replay probabilities $p_{s,i}$ and optimization weights $w_{s,i}$.
This auditing layer explicitly regulates how imagined experience enters policy optimization, complementing advances in simulator fidelity.

%% file: sec/component_ablation.tex
\subsection{Component Ablation}
\label{app:component_ablation}

The matched ablation retains the policy initialization, grounded budget, and refinement protocol.
StageTrust enables ratio/horizon scheduling; ChunkTrust enables reliability-based replay and loss weighting.
Their joint use gives the best results among these variants.

\begin{table}[t]
\centering
\small
\setlength{\tabcolsep}{5pt}
\caption{Component ablation on simulator evolution and two-level trust. Values preserve the original matched-control evaluation; three-seed external comparisons are reported separately in Tab.~\ref{tab:external_results}.}
\label{tab:ablation}
\begin{tabular}{lccc}
\toprule
Variant & Avg. SR $\uparrow$ & AUC $\uparrow$ & S2T-H $\downarrow$\\
\midrule
MFRL & 77.2 & 67.9 & 103.4\\
Static-MBRL & 73.3 & 59.8 & 168.9\\
Online-Updated MBRL & 76.2 & 65.4 & 139.6\\
Online-Updated+StageTrust & 77.9 & 68.4 & 113.2\\
Online-Updated+ChunkTrust & 77.5 & 67.6 & 121.5\\
\rowcolor{blue!8}
\MERL{} (Ours) & \textbf{79.7} & \textbf{70.6} & \textbf{92.1}\\
\bottomrule
\end{tabular}
\end{table}

%% file: sec/rebuttal_evidence.tex
\subsection{No-Oracle Trust Across Rollout Depths}
\label{app:no_oracle_validation}

This validation isolates the deployed residual predictor from access to matched future observations.
The predictor is frozen after calibration.
For recursive chunks at depths 2--4, matched future ground truth is withheld from scoring and used only afterward to evaluate residual estimates.
Depth counts action chunks from the grounded anchor, not individual environment steps.
The held-out set contains 600 depth-2 chunks and 900 depth-3--4 chunks, totaling 1500 chunks.
Intervals in Tab.~\ref{tab:depth_validation} are trajectory-bootstrap 95\% confidence intervals.

\begin{table}[t]
\centering
\small
\setlength{\tabcolsep}{4pt}
\caption{No-oracle trust estimation with matched future observations withheld from scoring. The first two rows give estimator reference results; subsequent rows report recursive-depth estimates with the corresponding trajectory-bootstrap 95\% confidence intervals.}
\label{tab:depth_validation}
\begin{tabular}{lcccc}
\toprule
Setting & Pearson $r$ $\uparrow$ & Rank-AUC $\uparrow$ & ECE $\downarrow$ & Top30 Prec. $\uparrow$\\
\midrule
Residual predictor & 0.71 & 0.82 & 0.063 & 0.79\\
Combined \MERL{} & 0.79 & 0.87 & 0.044 & 0.86\\
\midrule
Depth 2 ($N=600$) &
\shortstack{0.72\\{}[0.68,0.76]} &
\shortstack{0.82\\{}[0.79,0.85]} &
\shortstack{0.061\\{}[0.052,0.070]} &
\shortstack{0.80\\{}[0.76,0.84]}\\[0.5em]
Depths 3--4 ($N=900$) &
\shortstack{0.64\\{}[0.60,0.68]} &
\shortstack{0.77\\{}[0.74,0.80]} &
\shortstack{0.074\\{}[0.065,0.083]} &
\shortstack{0.74\\{}[0.70,0.78]}\\[0.5em]
All 2--4 ($N=1500$) &
\shortstack{0.68\\{}[0.65,0.71]} &
\shortstack{0.80\\{}[0.78,0.82]} &
\shortstack{0.067\\{}[0.060,0.074]} &
\shortstack{0.77\\{}[0.74,0.80]}\\
\bottomrule
\end{tabular}
\end{table}

Pearson $r$ measures association between predicted and observed residuals.
Rank-AUC measures ranking of lower-error chunks; ECE compares predicted trust with empirical reliability; Top30 precision is the fraction of truly low-error chunks among the top 30\% most trusted samples.
Prediction remains informative at depths 3--4, although ranking and calibration degrade relative to depth 2.
Predicted trust remains useful within the evaluated horizons, while grounded calibration and conservative stage-level scheduling are retained throughout refinement.

\paragraph{Downstream policy utility.}
We refine policies with matched numbers of chunks from Top, Mid, and Bottom trust bins, under the same update protocol and three training seeds.
Tab.~\ref{tab:trust_bin_utility} reports the resulting policy AUC.
The Top--Bottom contrast is $3.2\,[2.0,4.4]$ (95\% CI).
Matching chunk counts limits the explanation based on sample quantity alone; the experiment tests the utility of trust-ranked data under this controlled refinement protocol.

\begin{table}[t]
\centering
\small
\caption{Matched-count trust-bin policy AUC: mean $\pm$ SD over three seeds.}
\label{tab:trust_bin_utility}
\begin{tabular}{lccc}
\toprule
Trust bin & Top & Mid & Bottom\\
\midrule
Policy AUC $\uparrow$ & $69.9\pm0.8$ & $68.3\pm0.9$ & $66.7\pm1.0$\\
\bottomrule
\end{tabular}
\end{table}

\subsection{Hyperparameter Sensitivity}
\label{app:sensitivity}

We vary scheduler sharpness $\kappa$ and simulator proxy-loss weight $\lambda_q$ one at a time, using multipliers $\{0.5,1,2\}$ relative to their default settings.
The former changes the reliability-to-confidence mapping in Eq.~\eqref{eq:app_confidence_score}; the latter changes proxy supervision in Eq.~\eqref{eq:app_sim_loss}.
Across these settings, Avg. SR and AUC change by at most 1.4 and 1.6 points, respectively.
Every setting remains at least $+1.8$ Avg. SR and $+3.1$ AUC points above the controlled Online-Updated MBRL baseline.
The bounds apply to this one-at-a-time sweep and do not imply robustness to arbitrary joint changes.